\documentclass[11pt,a4paper]{article}

\usepackage[utf8]{inputenc}
\usepackage[T1]{fontenc}
\usepackage{lmodern}
\usepackage[margin=2.5cm,top=3cm,bottom=2.8cm,headheight=15pt]{geometry}
\usepackage{xcolor}
\usepackage{graphicx}
\usepackage{hyperref}
\usepackage{fancyhdr}
\usepackage{titlesec}
\usepackage{amsmath,amssymb}
\usepackage{booktabs}
\usepackage{caption}
\usepackage{enumitem}
\usepackage[numbers,sort&compress]{natbib}
\usepackage{tikz}
\usepackage{eso-pic}
\usepackage{tocloft}
\usepackage{lipsum}          
\usepackage{colortbl}
\usepackage{float}
\usepackage{microtype}
\usepackage{multirow}
\usepackage{array}
\usepackage{makecell}
\usepackage{framed}
\usepackage{longtable}
\usepackage{hhline}
\usepackage{fancyvrb}
\usepackage{fvextra}
\usepackage{multicol}
\usepackage{tablefootnote}
\usepackage{threeparttable}
\usepackage{tabularx}
\usepackage{mdframed}
\usepackage{subcaption}
\usepackage[usestackEOL]{stackengine}
\usepackage[colorinlistoftodos]{todonotes}
\usepackage{cleveref}

\definecolor{MeshyLime}{HTML}{C5F955}
\definecolor{MeshyLimeDark}{HTML}{7DA82A}       
\definecolor{MeshyBlack}{HTML}{181818}
\definecolor{MeshyPink}{HTML}{FF3E8F}
\definecolor{MeshyDarkGray}{HTML}{2A2A2A}
\definecolor{MeshyMidGray}{HTML}{666666}
\definecolor{MeshyLightGray}{HTML}{F5F5F5}
\definecolor{MeshyAccentLine}{HTML}{B8E86E}     

\hypersetup{
    colorlinks=true,
    linkcolor=MeshyDarkGray,
    citecolor=MeshyLimeDark,
    urlcolor=MeshyPink,
    pdfauthor={Meshy AI},
    pdftitle={CLOVER: Exploring Unified Text-Conditioned Image and 3D Generation},
}



\RequirePackage{xspace}
\makeatletter
\DeclareRobustCommand\onedot{\futurelet\@let@token\@onedot}
\def\@onedot{\ifx\@let@token.\else.\null\fi\xspace}

\makeatother

\newcommand{\mymodel}{Omni123\xspace}
\newcommand{\methodname}{Omni123\xspace}

\renewcommand{\paragraph}[1]{\noindent\textbf{#1.}\hspace*{1em}}

\titleformat{\section}
  {\Large\bfseries\color{MeshyBlack}}
  {\color{MeshyLimeDark}\thesection}{0.6em}{}

\titleformat{\subsection}
  {\large\bfseries\color{MeshyBlack}}
  {\color{MeshyLimeDark}\thesubsection}{0.5em}{}

\titleformat{\subsubsection}
  {\normalsize\bfseries\color{MeshyDarkGray}}
  {\color{MeshyPink!70}\thesubsubsection}{0.5em}{}

\titlespacing*{\section}{0pt}{2.0ex plus 0.6ex}{1.0ex plus 0.3ex}
\titlespacing*{\subsection}{0pt}{1.5ex plus 0.4ex}{0.7ex plus 0.2ex}
\titlespacing*{\subsubsection}{0pt}{1.2ex plus 0.3ex}{0.5ex plus 0.1ex}

\fancypagestyle{titlepage}{%
  \fancyhf{}%
  \fancyfoot[C]{%
    {\color{MeshyLime}\rule{12pt}{1.5pt}}\enspace
    {\small\color{MeshyMidGray}\thepage}\enspace
    {\color{MeshyLime}\rule{12pt}{1.5pt}}%
  }%
}

\newcommand{\meshyaccentbar}{%
  \AddToShipoutPictureBG*{%
    \AtPageUpperLeft{%
      \raisebox{-3.5pt}{%
        \color{MeshyLime}\rule{\paperwidth}{3.5pt}%
      }%
    }%
  }%
}

\setlist[itemize]{
  leftmargin=15pt, nosep,
  label={\color{MeshyLime}\textbullet},
}
\setlist[enumerate]{
  leftmargin=15pt, nosep,
  label={\color{MeshyBlack}\arabic*.},
}

\newcommand{\email}[1]{\href{mailto:#1}{\texttt{#1}}}

\begin{document}

\meshyaccentbar
\thispagestyle{titlepage}
\vspace*{-16mm}

\noindent
\includegraphics[height=1.0cm]{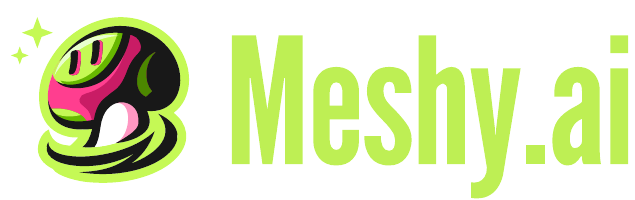}%


\noindent{\color{MeshyAccentLine}\rule{\textwidth}{1pt}}



\begin{center}
  \includegraphics[height=2.0cm,width=7.0cm]{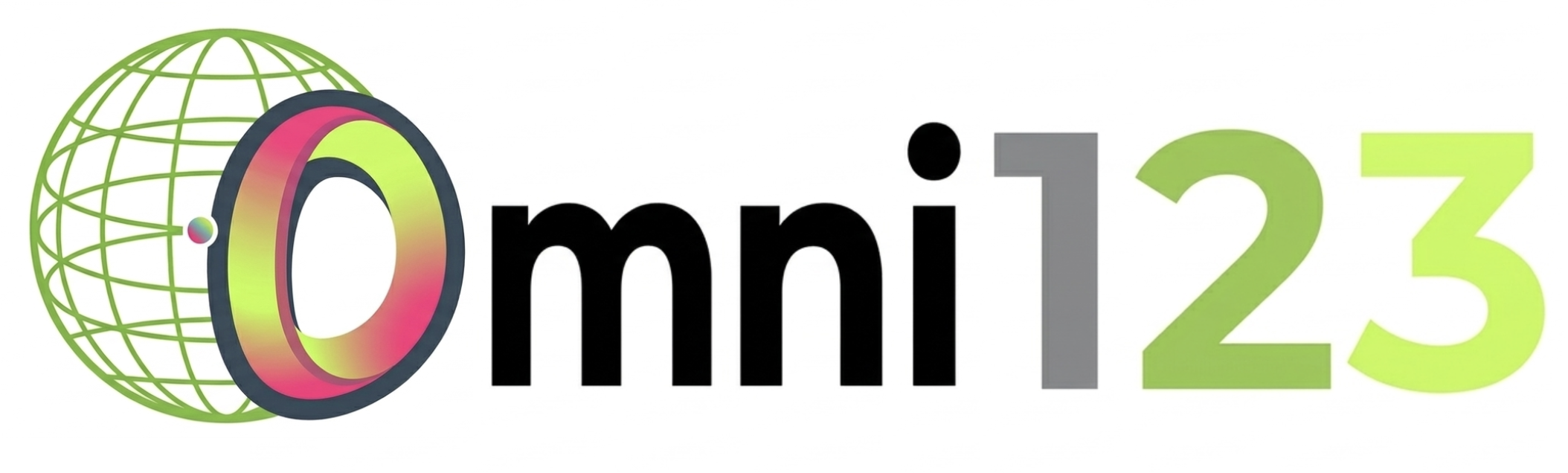}%

  \vspace{1mm}

   {\fontsize{22}{22}\selectfont\bfseries\color{MeshyBlack}%
    Exploring 3D Native Foundation Models\\
    with Limited 3D Data by Unifying\\
    Text to 2D and 3D Generation%
  }
\end{center}

\vspace{1mm}

\noindent{\color{MeshyAccentLine}\rule{\textwidth}{1pt}}

\vspace{1mm}

\begin{center}
  {\normalsize\color{MeshyMidGray}%
    Chongjie Ye\textsuperscript{1,2}\,,\quad
    Cheng Cao\textsuperscript{3}\,,\quad
    Chuanyu Pan\textsuperscript{3}\,,\quad
    Yiming Hao\textsuperscript{2}\,,\quad
    Yihao Zhi\textsuperscript{2}\,,\quad \\
    Yuanming Hu\textsuperscript{3}\,,\quad
    Xiaoguang Han\textsuperscript{2,1,*}%
  }\\
  \vspace{2mm}
  {\small\color{MeshyMidGray}%
    \textsuperscript{1}FNii-Shenzhen \quad
    \textsuperscript{2}SSE, CUHK(SZ) \quad
    \textsuperscript{3}Meshy AI%
  }\\
  \vspace{1mm}
  {\small\color{MeshyMidGray}%
    \textsuperscript{*}Corresponding author:\;
    \email{hanxiaoguang@cuhk.edu.cn}%
  }
\end{center}
\vspace{1mm}

\noindent
\begin{tikzpicture}
  \node[
    draw=MeshyAccentLine, fill=white,
    rounded corners=5pt, line width=0.8pt,
    inner sep=14pt,
    text width=\dimexpr\textwidth-30pt,
  ] {%
    \noindent{\normalsize\bfseries Abstract}\par\vspace{0.2em}

     \footnotesize
    Recent multimodal large language models have achieved remarkable success in unified text and image understanding and generation, yet extending such native capability to 3D remains an open challenge. The core bottleneck is data: compared to the near-infinite 2D imagery on the web, high-quality 3D assets are orders of magnitude scarcer, leaving 3D synthesis severely under-constrained. Current methods often circumvent this limitation through indirect pipelines that edit in 2D image space and lift results into 3D via iterative optimization, sacrificing geometric consistency and the straightforwardness of native generation. We present Omni123, a 3D native foundation model that addresses limited 3D data by unifying text-to-2D and text-to-3D generation within a single autoregressive framework. Our key insight is that cross-modal generative consistency between images and 3D can act as an implicit structural constraint: by representing text, images, and 3D geometry as discrete tokens in a shared sequence space, the model can leverage abundant 2D observations as a rich geometric prior to fortify 3D representations. To realize this, we introduce an interleaved X-to-X training paradigm that coordinates diverse cross-modal tasks over heterogeneous paired datasets without requiring fully aligned text–image–3D triplets. By traversing semantic–visual–geometric cycles (e.g., text → image → 3D → image) within single autoregressive sequences, the model learns representations that simultaneously satisfy high-level semantic intent, appearance fidelity, and multi-view geometric consistency, while mitigating harmful interference between appearance and geometry objectives. Extensive experiments demonstrate that Omni123 significantly improves geometric consistency and semantic alignment in text-guided 3D generation and editing, validating that unifying 2D and 3D generative processes provides an effective and scalable pathway toward multimodal 3D world models.


  };
\end{tikzpicture}

\vspace{4mm}
\begin{figure}[H]
  \centering
  \includegraphics[width=1\textwidth]{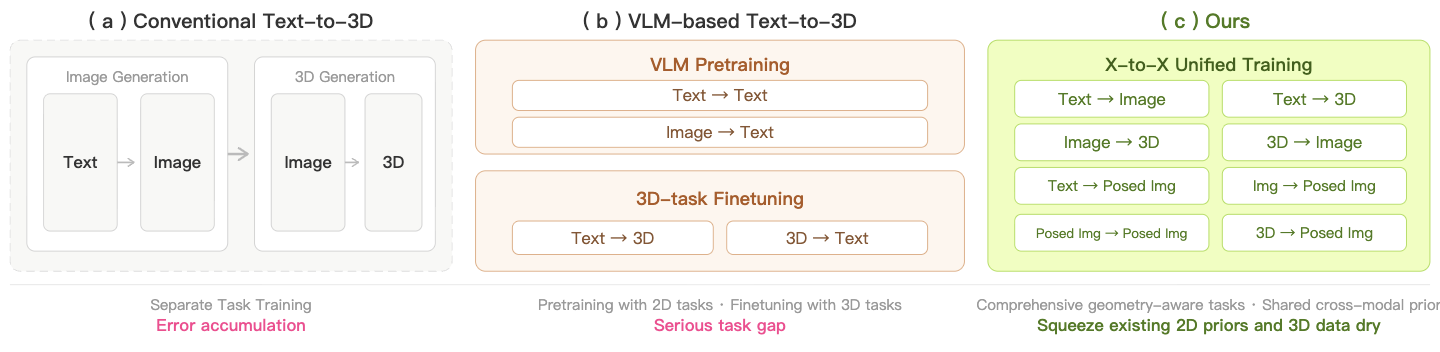}
  \caption{\textbf{Comparison of Text-to-3D paradigms} }
  \label{fig:teaser_small}
\end{figure}


\begin{figure*}[t]
  \centering
  \includegraphics[width=\textwidth]{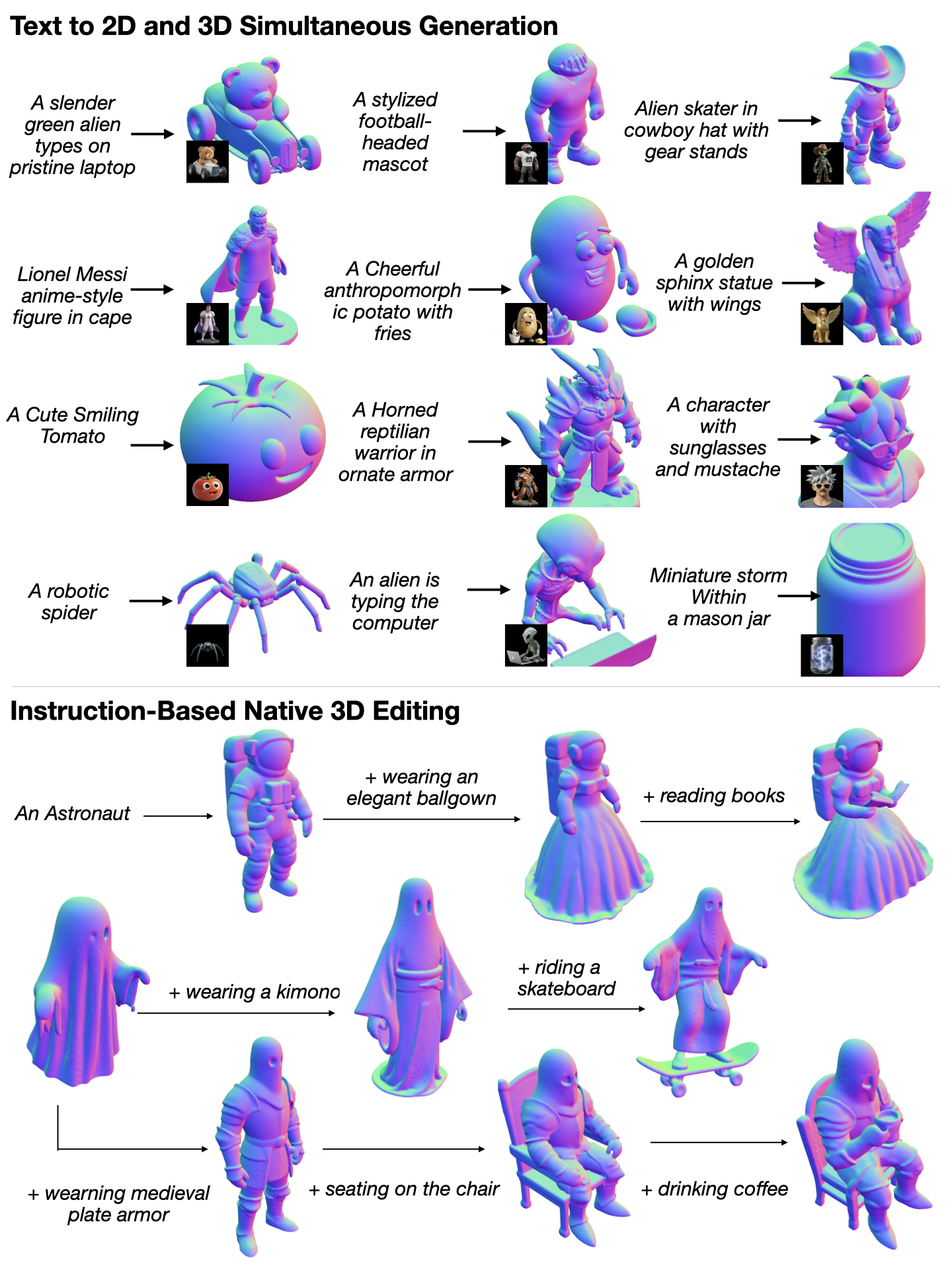}
  \caption{\textbf{\mymodel} enables native 3D generation and editing within a unified multimodal framework trained with limited 3D data. Top: Text-to-joint-2D-and-3D generation results across diverse prompts. Each example shows the generated 2D image (inset) alongside the normal map of the corresponding 3D output, demonstrating high geometric fidelity and strong semantic alignment with the input text. Bottom: Native 3D editing of 3D models through sequential and branching text instructions (e.g., ``+wearing a kimono,'' ``+riding a skateboard''), illustrating the model's ability to perform iterative and diverse 3D modifications directly in 3D space. More qualitative results are shown in Figures~\ref{fig:showcase_1} and~\ref{fig:showcase_2}.}
  \label{fig:teaser}
\end{figure*}

\newpage
\tableofcontents
\newpage


\section{Introduction}
\label{sec:intro}

Recent breakthroughs in generative AI have catalyzed a shift toward unified multimodal models capable of jointly reasoning across linguistic and visual domains, with models like GPT-4o\cite{openai_gpt_image_1_2025} and Nano-Banana~\cite{google_nanobanana_gemini_2026} enabling native capabilities for high-fidelity synthesis and instruction-guided editing. As the community pushes toward more immersive digital environments, the natural evolution is the quest for a "3D Nano-Banana": a unified framework capable of native 3D generation and editing, which is essential for advancing applications in embodied AI and autonomous virtual world synthesis. However, achieving such native capability in 3D remains a persistent challenge.

Unlike 2D AIGC, which leverages near-infinite web-scale data, 3D generation faces acute data scarcity. This disparity leaves 3D synthesis inherently under-constrained, especially under sparse supervision from text or single views. To compensate, current methods often treat 2D images as an intermediate proxy—performing edits in the image space and lifting them into 3D through iterative optimization. While effective, these indirect approaches frequently compromise geometric consistency and lack the straightforwardness of 2D workflows. This suggests that 3D learning requires stronger structural constraints than sparse signals can provide. Crucially, as 2D visual data implicitly encodes deep 3D structural information, including object shape~\cite{lin2025depth}, appearance~\cite{wang2025mage}, and spatial relationships~\cite{baik2025learning}, it prompts a pivotal hypothesis: Can we leverage the abundance of 2D observations to fortify 3D generation via a unified multimodal framework?

A common strategy involves aggregating multiple cross-modal tasks. However, it does not inherently guarantee effective knowledge transfer. This stems from the fact that different generative tasks are rooted in distinct statistical priors: the text-to-image task focuses on modeling appearance distributions, whereas the text-to-3D task requires consistent geometric reasoning.  Without meticulous coordination, joint training may introduce both \textbf{beneficial transfer and harmful task interference}, where competing gradients may ultimately degrade the quality of the learned 3D representations.

To address this challenge, we introduce an \textbf{interleaved X-to-X training paradigm}. \textbf{We posit that cross-modal generative consistency between images and 3D can act as an implicit structural constraint for learning robust 3D representations.} Different modalities offer complementary observations of the same underlying object: text describes high-level semantic intent, images capture rich appearance cues and partial geometric hints, while 3D representations provide explicit spatial structure and multi-view consistency. Consequently, we propose that cross-modal generative consistency between images and 3D can serve as a potent implicit constraint. Specifically, by forcing a model to traverse a "semantic-visual-geometric" cycle (e.g., text $\rightarrow$ image $\rightarrow$ 3D $\rightarrow$ image), the resulting representations could simultaneously satisfy high-level intent, appearance fidelity, and multi-view consistency. 

We investigate this hypothesis through a unified autoregressive formulation that treats text, image, and 3D geometry as discrete tokens within a shared sequence space. To circumvent the reliance on fully paired text–image–3D triplets, we leverage heterogeneous paired datasets. The model is trained on a diverse set of conditional generation tasks where any modality can serve as input, and the target modality is predicted autoregressively. This formulation maximizes the utility of available multimodal data and ensures that the model does not merely learn isolated tasks, but rather develops a unified latent representation across the semantic, visual, and geometric domains.

In summary, we incorporate the following key technical advancements:

\begin{itemize}
    \item We introduce a framework that tokenizes text, images, and 3D geometry within a shared sequence space. This enables native 3D reasoning and editing, eliminating the reliance on intermediate 2D proxies and iterative image-space optimization.
    \item We propose an interleaved training scheme that coordinates diverse generative tasks across heterogeneous datasets. This approach mitigates task interference between appearance and geometry priors, facilitating beneficial cross-modal knowledge transfer from 2D to 3D.
    \item We leverage generative consistency across "semantic-visual-geometric" cycles as an implicit structural constraint. This allows abundant 2D observations to serve as a rich geometric prior, effectively fortifying 3D representations against acute data scarcity.

\end{itemize}

Through extensive experiments, we study how different multimodal objectives interact during training and how cross-modal generative processes influence the quality of learned 3D representations. Our results indicate that interleaved cross-modal generation significantly improves geometric consistency and semantic alignment in text-guided 3D generation and editing tasks. These findings suggest that \textbf{cross-modal generative consistency provides an effective pathway toward scalable multimodal 3D world models.}






\section{Related Work}
\label{sec:related work}
\subsection{Image Generation \& Editing}
Generative adversarial network (GAN) is the first representative image generation~\cite{goodfellow2014generative,brock2018large,karras2019style,kang2023scaling} and editing~\cite{pan2023drag,ling2024freedrag} method in the deep learning era. To improve the distribution coverage, subsequent research also explored likelihood-based generative models. Diffusion models~\cite{ho2020denoising,dhariwal2021diffusion,rombach2022high,podell2023sdxl,chen2023pixart} and Flow Matching~\cite{lipman2022flow,flux2024,esser2024scaling} later became the dominant paradigm by formulating image synthesis as the reverse denoising process from noise to data. Beyond their strong generation quality, diffusion models provide a more expressive latent space and a natural interface for conditioning, which has driven the rapid development of image editing methods, including controllable generation\cite{zhang2023adding,tan2025ominicontrol}, inversion-based training-free editing\cite{dalva2024fluxspace,fan2024videoshop}, and optimization-based editing\cite{ren2025fds,jeong2024dreammotion}. More recently, emerging large-scale visual generation models\cite{wu2025qwen,flux-2-2025,cai2025z} have increasingly explored leveraging stronger multi-modal reasoning and instruction-following capabilities of VLMs to unify image editing, generation, and understanding.

\subsection{3D Generation \& Editing}
3D generation has evolved along two main paradigms. \cite{liu2023one,qian2023magic123} relies on pretrained image/video diffusion models and distills their priors into 3D representations through score distillation. While effective, this indirect 2D supervision often leads to slow optimization, inconsistent 3D appearance, and artifacts. In contrast, recent native 3D generative models \cite{xiang2025structured,xiang2025native,hunyuan3d-2.1} learns 3D distribution directly, providing a more efficient and scalable route toward high-quality 3D synthesis and already seeing broad adoption in commercial applications\cite{lai2025hunyuan3d-2.5,hyper3d_rodin_2026,tripo3d_2026,meshy_2026}. Building on these strong foundation models, many recent works \cite{hunyuan3d2025hunyuan3d-omni,ye2025nano3d,hu2026easy3e} have further explored editing tasks. Nevertheless, direct end-to-end text-3D synthesis \cite{poole2022dreamfusion,hu2025turbo3d,qiu2024richdreamer} still lags far behind due to the large modality gap. To address this challenge, we propose a text-image-3D-image interleaved training paradigm that uses images as a bridge to alleviate both the modality gap and the conflict between generation and reconstruction, achieving state-of-the-art text-3D performance.

\subsection{Multimodal Generation}  
Numerous previous works have explored extending LLMs toward visual generation and understanding via mapping learned visual tokens into the text feature space of a pretrained diffusion~\cite{koh2023generating,sun2023emu,cruz2023scaling}. Later studies further explore an end-to-end unified architecture for generation and understanding across modalities~\cite{deng2505emerging,xie2025show,wang2024emu3,liu2025tuna,chen2025janus}. More recently, GPT-4o~\cite{openai_gpt_image_1_2025} and Nano-Banana~\cite{google_nanobanana_gemini_2026} have further accelerated this progress.

\noindent 
For the 3D modality, ShapeLLM-Omni~\cite{ye2025shapellm} is among the first to unify 3D generation and understanding within a single autoregressive framework. However, it purely suffers from quantization-induced information loss, which limits generation quality. Lucid3D~\cite{chenlucid} further extends to 3D editing, while Cube3D\cite{bhat2025cube} improves the 3D tokenizer with self-supervised objectives, leading to higher fidelity and further extending to 3D scene generation. AR3D-R1\cite{tang2025we} introduces the first 3D RL system to improve 3D consistency and alignment with human preferences. However, due to the scarcity and expense of 3D data, the scalability and generalization of previous works have been constrained. To alleviate this limitation, we unify 2D and 3D generation so that 3D tasks can benefit from rich and general visual knowledge from large-scale 2D data. This enables high-quality 3D creation with only limited text-image-3D pairs, while also yielding emergent generation and editing capabilities.




\section{Data Pre-Processing}
\label{sec:data}
 
In this section, we describe the composition of our multi-modal dataset, the cleaning and filtering pipelines applied to each modality, the data synthesis pipeline used to scale high-quality SFT data, and the captioning strategies used to bridge modalities.
 
\subsection{Data Overview}
\label{sec:data:overview}
 
Our training corpus spans three modalities---text, images, and 3D assets---and is 
organized into the following paired and triplet subsets. Table~\ref{tab:data_stats} 
summarizes the complete data composition across pre-training and supervised 
fine-tuning stages.

\begin{table}[h]
\centering
\caption{Summary of training data composition. Token counts per sample are 
computed using 77 text tokens, 1{,}024 image tokens per view, and 1{,}024 3D mesh 
tokens. The table is divided into pairwise pre-training data (upper) and 
continued training \& supervised fine-tuning data (lower). Total tokens shown 
are the dataset size; tokens seen during training are higher due to multiple epochs 
(see Table~\ref{tab:training_recipe}).}
\label{tab:data_stats}
\setlength{\tabcolsep}{8pt}
\renewcommand{\arraystretch}{1.15}
\resizebox{\linewidth}{!}{
\begin{tabular}{clcccr}
\toprule
\textbf{Stage} & \textbf{Data Configuration} & \textbf{Items} & \textbf{Tokens/Sample} & \textbf{Total Tokens} & \textbf{Token \%} \\
\midrule
\multirow{4}{*}{Pre-Training}
& Text + Image           & 63.7M  & 1,101 & 70.1B  & 14.8\% \\
& Image + 3D             & 120M  & 2,048 & 245.8B & 51.8\% \\
& Text + 3D              & 9M   & 1,101 & 9.9B   & 2.1\%  \\
\cmidrule(lr){2-6}
& \textit{Subtotal}      & \textit{192.7M} & \textit{---} & \textit{325.8B} & \textit{68.6\%} \\
\midrule
\multirow{3}{*}{CT \& SFT}
& Text + Single-View Image + 3D    & 2M   & 2,125 & 4.3B   & 0.9\%  \\
& Text + 6 Posed Images + 3D       & 20M  & 7,245 & 144.9B & 30.5\% \\
\cmidrule(lr){2-6}
& \textit{Subtotal}                 & \textit{22M}  & \textit{---} & \textit{149.2B} & \textit{31.4\%} \\
\midrule
& \textbf{Total}                    & \textbf{214.7M} & \textbf{---} & \textbf{475.0B} & \textbf{100\%} \\
\bottomrule
\end{tabular}
}
\end{table}
 
The pre-training stage uses 194.7\,M pairwise samples spanning three cross-modal 
directions (Text$\leftrightarrow$Image, Image$\leftrightarrow$3D, 
Text$\leftrightarrow$3D), totaling 328.0\,B tokens in the dataset itself. 
The interleaved SFT stage introduces two chain-of-generation tasks: 
(i)~a single-view pipeline (Text$\to$Image$\to$3D) with 2\,M triplets, and 
(ii)~a multi-view pipeline (Text$\to$6 Posed Images$\to$3D) with 20\,M samples, 
where six views are rendered from known camera poses. Despite comprising only 
9.2\% of all items, the multi-view SFT data accounts for 30.4\% of all tokens 
due to the much longer per-sample sequence length (7{,}245 tokens). Note that 
the total tokens seen during training (reported in Table~\ref{tab:training_recipe}) 
is higher (${\sim}1.16$\,T for pre-training) due to multiple epochs over the data. 
Table~\ref{tab:tokenization} details the tokenization configuration and sequence 
lengths for each modality.


\begin{table}[t]
\centering
\caption{Per-modality tokenization configuration.}
\label{tab:tokenization}
\setlength{\tabcolsep}{8pt}
\renewcommand{\arraystretch}{1.12}
\begin{tabular}{@{}lcp{8.8cm}@{}}
\toprule
\textbf{Modality} & \textbf{Tokens} & \textbf{Tokenizer} \\
\midrule
Text 
& 77 
& CLIP~\cite{radford2021learning} text encoder + Qwen3-0.6B~\cite{yang2025qwen3} for enhanced semantic understanding \\

Image 
& 1{,}024 / view 
& Our image tokenizer, a 1D discrete visual tokenizer with semantic regularization \\

3D Mesh 
& 1{,}024 
& Cube3D~\cite{bhat2025cube}, a Perceiver-based VQ-VAE with optimal-transport vector quantization \\
\bottomrule
\end{tabular}
\end{table}
 
\subsection{Text--Image Pairs}
\label{sec:data:text_image}
 
The text--image pre-training corpus contains 63.7\,M open-domain images in total, comprising proprietary data mixed with synthetic data by Z-Image~\cite{cai2025z}.

A natural question is whether the 120\,M rendered images from our Image--3D 
corpus (Section~\ref{sec:data:3d_processing}) should also be included, given 
their scale. We deliberately choose \emph{not} to do so: these renderings 
exhibit uniform lighting, synthetic material appearance, and the absence of 
natural backgrounds, diverging substantially from in-the-wild photographs. 
Mixing the two domains in the text--image objective would create a distribution 
conflict that degrades image generation quality. We therefore confine rendered 
imagery to the Image$\to$3D and Text$\to$3D objectives, where the rendering 
distribution is consistent with the target task.
 
\subsection{Image--3D Pairs}
\label{sec:data:3d_processing}
 
The image--3D pairs constitute the largest subset (120\,M) and are critical for 
learning the reconstruction mapping between 2D observations and 3D geometry. 
Following established practices in recent large-scale 3D data 
engineering~\cite{hy3dbench,triposg,clay,step1x3d}, we process this corpus 
through a three-stage pipeline:

(1)~\textit{Rendering and format conversion}: 3D assets from heterogeneous sources are converted into meshes, aligned to a canonical orientation, standardized with PBR textures, saved as GLB files, and rendered using Meshy's in-house renderer with HDRI environments and lighting conditions sampled from a pool of 2{,}000 setups.

(2)~\textit{Asset filtering}: we exclude assets with poor geometric quality 
(low polygon counts, simplistic topology), poor texture quality (UV defects, 
low resolution), noisy photogrammetry scans, and large thin-walled structures---the latter causing SDF sign 
discontinuities that destabilize training and multi-view inconsistencies where thin geometry becomes invisible under certain viewpoints.

(3)~\textit{Post-processing}: filtered meshes undergo watertight conversion, narrow-band SDF sampling, and point cloud sampling.

\subsection{Text--3D Pairs}
\label{sec:data:captioning}
 
Existing captions for 3D objects~\cite{cap3d} either suffer from poor alignment 
with the geometry they describe or lack sufficient detail, which limits 
high-quality text-to-3D generation. We therefore design a multi-granularity 
captioning pipeline utilizing visual chain-of-thought reasoning, to produce multiple precise and detailed text descriptions focusing on different aspects of a 3D model. When applicable, metadata is also supplied to the model for factual world knowledge grounding.

The pipeline is run on each of the 5\,M filtered 3D assets, consisting of three stages:

(1)~\textit{Visual Chain of Thought Analysis}: we feed multi-view rendered images of each asset into a Visual-Language-Model (VLM) such as Gemini-3~\cite{gemini3} to generate highly detailed per-view descriptions that are used as anchors for the model to perform orientation and spatial correspondence analysis. After the model has identified spatial correspondence between images and therefore oriented the object, the VLM is asked to analyze the model in terms of its appearance, geometry, potential function, and origin in the real world.

(2)~\textit{Captioning and Categorization}: The VLM takes the analysis and produces a paragraph-long detailed caption that goes over all aspects of the 3D model. Categories are then assigned by the VLM, producing rough semantic classification.

(3)~\textit{Human captions mimicking}: We prepared 10K human-labeled short captions distributed over all categories, and 4 random examples are selected per-asset based on its evaluated category to perform few-shot caption augmentation. Human captions leveraged in this way greatly improve the lexical and syntactical diversity of our captions.

All three stages are completed within one continuous multi-turn conversation, preserving all context to the model --- including the images. Reasoning is enabled, and the reasoning framework is enforced through schema-guided generation.

Through this multi-granularity captioning process, each of the 5\,M filtered 3D assets is augmented with multiple caption variants. We then perform an alignment filtering step, removing pairs where the caption poorly matches the 3D geometry based on VLM-evaluated consistency scores. This yields a total of 9\,M Text--3D training pairs used in pre-training (Section~\ref{sec:model:pretraining}).

\subsection{Interleaved SFT Triplets}
\label{sec:data:sft_triplets}
 
While the Objaverse corpus~\cite{objaverse,objaversexl} provides a substantial 
volume of image--3D training pairs, the majority of its assets exhibit simple 
structures and plain surface details, limiting the generation capabilities of 
models trained exclusively on this data. Given the prohibitive cost of manually creating high-quality 3D assets, we construct a data synthesis pipeline that 
performs Text$\to$Image$\to$3D generation, producing a dataset of synthesized 3D assets with considerably 
more complex structures and richer geometric detail.
 
\textit{Step 1: Prompt Curation.}
We initiate the synthesis process from text prompts rather than image prompts, 
as text affords more direct control over semantic diversity and ensures variety 
in the resulting geometries. We first source approximately 14\,M high-quality 
raw prompts from DiffusionDB~\cite{diffusiondb}. A Gemini-3-Flash~\cite{gemini3} classifier is applied to filter out complex multi-object scenes, 
retaining only single-object descriptions amenable to 3D asset generation. 
Rule-based filtering further eliminates stylistic modifiers (e.g., 
``cinematic'', ``4k wallpaper''), followed by a structural standardization 
step to enforce consistent prompt formatting. After all filtering stages, 
approximately 2\,M prompts are retained.
 
\textit{Step 2: Text-to-Image Generation.}
We employ Z-Image-Turbo\cite{cai2025z} for high-throughput text-to-image generation. 
The text prompt conditioning is augmented with viewpoint and lighting 
specifications to produce object-centric images with controlled camera angles, 
which is important for stable downstream 3D reconstruction.

\textit{Step 3: Background Removal.}
To obtain clean object masks suitable for image-to-3D generation, we apply 
BiRefNet~\cite{birefnet} for high-resolution foreground segmentation. 
We then subsequently use VLM to
evaluates each masked image for segmentation artifacts (e.g., residual 
background, truncated object parts) and filters out samples with poor mask 
quality. It also simultaneously re-captions the masked object to 
produce an aligned text description, ensuring cross-modal consistency between 
the final caption and the visual content.
 
\textit{Step 4: Image-to-3D Synthesis.}
We adopt Meshy 4 to synthesize 3D meshes from the masked images. The generated 
meshes undergo the same watertight processing and point cloud sampling 
pipeline described in Section~\ref{sec:data:3d_processing}.
 
\textit{Step 5: Quality Filtering.}
A rigorous data cleaning process combines automated geometric assessment 
with MLLM-based evaluation. Specifically, we render multi-view images of each synthesized mesh and evaluate three quality axes: 
(i)~\textit{geometric consistency}---whether the object is structurally 
sound with preserved symmetry and no warped geometry; 
(ii)~\textit{surface texture clarity}---whether micro-details (e.g., wood 
grain, metal finish, fabric weave) are sharp rather than flat or muddy; and 
(iii)~\textit{silhouette connectivity}---whether the object has a clear, 
recognizable silhouette without heavy occlusion or detached floaters. 
For each axis, the MLLM assigns a binary ``high''/``low'' rating, and only samples rated ``high'' on all three axes are retained in the final dataset.
 
\textit{Multi-Granularity Re-Captioning.}
Each retained asset is re-captioned at three levels of specificity by an MLLM: 
a \textit{long} caption (40--70 words) providing specification-level detail 
including materials, textures, colors, and geometric complexity; a 
\textit{medium} caption (10--25 words) covering subject, primary material, 
and basic geometry; and a \textit{short} caption (1--5 words) capturing only 
the core subject identity. These multi-granularity captions serve both the 
SFT triplet task (Text$\to$Image$\to$3D) and the multi-view SFT task 
(Text$\to$6~Posed~Images$\to$3D).
 
\section{Model Training}
\label{sec:method}

\subsection{Architecture}
\label{sec:model}
A key obstacle to unified multi-modal generation is the heterogeneity of representations across modalities. Text is naturally a sequence of discrete symbols; images are typically encoded as 2D spatial grids of continuous features~\cite{rombach2022high,agarwal2025cosmos}; and 3D shapes are represented as volumetric lattices~\cite{ye2025shapellm}, unordered point sets, or continuous latent vectors~\cite{zhang20233dshape2vecset}. These representations are incompatible with autoregressive language modeling in two fundamental ways. First, \emph{continuous} latent spaces---such as the latent vectors produced by 3DShape2VecSet~\cite{zhang20233dshape2vecset}---lack the discrete vocabulary required for next-token prediction;
Second, even when discretized, \emph{2D or 3D spatial} tokenizers produce token counts that scale with spatial resolution, yielding intractable context lengths, and their modality-specific positional encodings and attention patterns fragment the model and hinder cross-modal knowledge transfer.

Building on recent advances in \emph{vector-quantized} tokenization~\cite{bhat2025cube,gigatok} that convert continuous representations into discrete codebook indices amenable to next-token prediction, and \emph{1D} tokenization that compresses each modality into a short, structurally uniform sequence, \methodname{} adopts a self-developed representative image tokenizer (Section~\ref{sec:pretraining:image_tokenizer}) for images and Cube3D~\cite{bhat2025cube} geometry tokenizer for 3D meshes. This combination yields compact 1D discrete token sequences for both modalities, enabling seamless concatenation without modality-specific positional encoding. Both modalities are then processed by a shared autoregressive Transformer backbone that maximizes parameter sharing and enforces cross-modal coherence through shared self-attention.

\begin{figure*}[t]
\centering
\includegraphics[width=\textwidth]{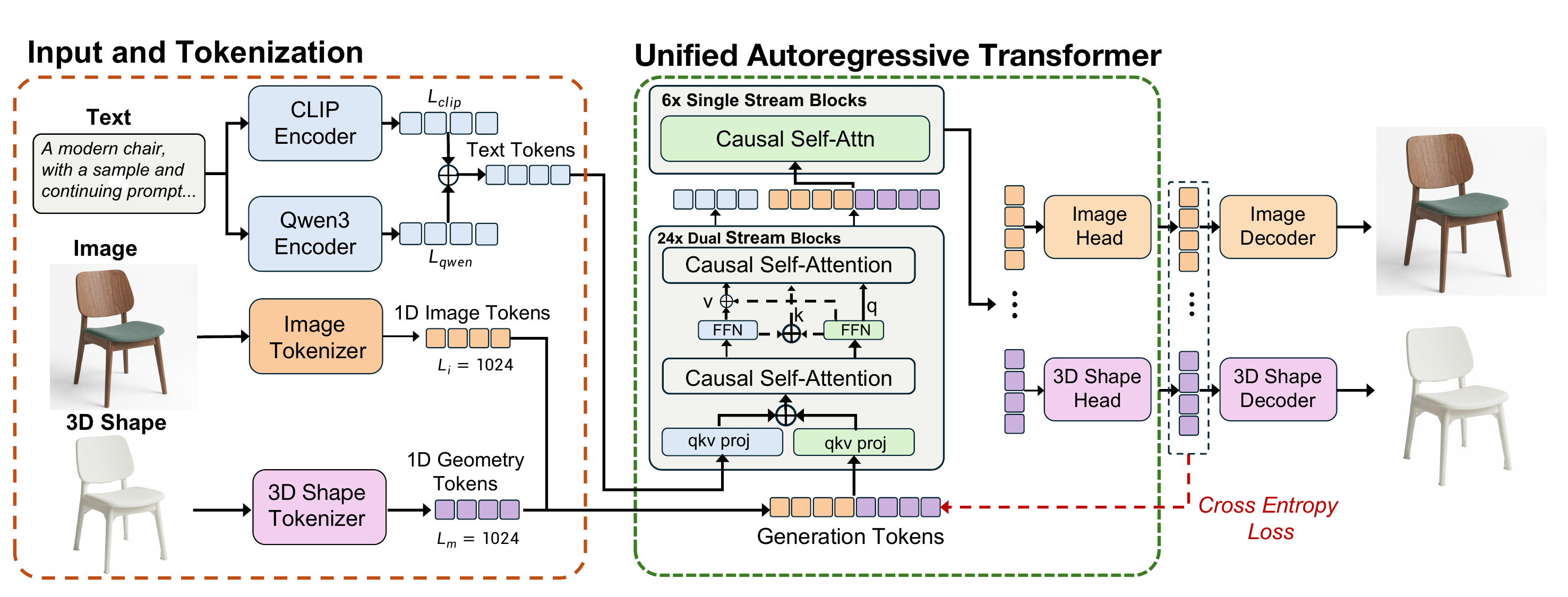}
\caption{Overview of the \textbf{\methodname{}} architecture. Text is encoded by dual text encoders (CLIP~\cite{radford2021learning} and Qwen3-0.6B~\cite{yang2025qwen3}) and fed into a conditioning stream, while images and 3D shapes are tokenized into 1D discrete tokens and concatenated into a unified generation stream. The unified autoregressive 
transformer backbone uses 24 dual-stream blocks to jointly process the conditioning and generation tokens under causal attention, followed by 6 single-stream layers operating only on generation tokens, and finally with modality-specific linear heads decoding token logits over the 2D and 3D codebooks.}
\label{fig:architecture}
\end{figure*}

As illustrated in Figure~\ref{fig:architecture}, the architecture consists of modality-specific tokenizers, a unified autoregressive Transformer backbone, and modality-specific output heads. For text conditioning, we employ two complementary encoders: CLIP~\cite{radford2021learning} provides $L_{\text{clip}} = 77$ vision-language-aligned embeddings, while Qwen3-0.6B~\cite{yang2025qwen3} provides $L_{\text{qwen}}$ embeddings that capture richer linguistic detail; the two are concatenated and injected via cross-attention. Images are tokenized into $L_i = 1{,}024$ 1D tokens via our image tokenizer; and 3D geometry are encoded into $L_m = 1{,}024$ 1D \emph{geometry tokens} via Cube3D~\cite{bhat2025cube} shape tokenizer (detailed in Sections~\ref{sec:model:3d_tokenizer} and~\ref{sec:pretraining:image_tokenizer}). Image and 3D shape tokens are concatenated at the sequence level and processed by the backbone autoregressively.

The backbone follows the dual-stream autoregressive architecture of Cube3D~\cite{bhat2025cube}. A \emph{conditioning stream} carrying text embeddings and a \emph{generation stream} carrying image and 3D shape tokens run in parallel through the dual-stream layers. At each dual-stream layer, the two streams are independently projected into queries, keys, and values; these are then concatenated along the sequence dimension and processed by a single joint attention operation under a causal mask. The attention output is split back along the sequence boundary and used to update both streams through separate feed-forward sub-layers. In the final dual-stream layer, the conditioning stream contributes only keys and values---its query projection and post-attention update are omitted, so the information flow becomes unidirectional from conditioning to generation. The subsequent single-stream layers retain only causal self-attention over the generation tokens, without further text interaction. Within the generation stream, image and 3D shape tokens are concatenated into a single flat sequence and processed by fully shared self-attention weights---no modality-specific branches or attention masks are introduced. This means every attention layer implicitly performs cross-modal fusion, and all generation-side parameters are shared across modalities, enabling visual priors learned from the large-scale Text$\to$Image data to directly benefit the data-scarcer Text$\to$3D task. Each transformer layer employs SwiGLU~\cite{shazeer2020glu} as the feed-forward network with a $4\times$ expansion ratio, following the design of Cube3D~\cite{cube}. Detailed architecture hyperparameters are listed in Table~\ref{tab:model_config}.

\begin{table}[t]
\centering
\caption{Model architecture configuration.}
\label{tab:model_config}
\begin{tabular}{@{}lc@{}}
\toprule
\textbf{Hyperparameter} & \textbf{Value} \\
\midrule
Dual-stream (joint-attention) layers & 24 \\
Single-stream (self-attention) layers & 6 \\
Hidden dimension $d$ & 1{,}536 \\
Attention heads & 12 \\
Per-head dimension & 128 \\
FFN expansion ratio & $4\times$ (SwiGLU) \\
Text encoder (vision-language) & CLIP ($L_{\text{clip}} = 77$) \\
Text encoder (linguistic) & Qwen3-0.6B \\
Image tokens $L_i$ & 1{,}024  \\
3D shape tokens $L_m$ & 1{,}024 \\
3D codebook embedding dim & 16 \\
RoPE $\theta$ & 1{,}000 \\
Layer norm type & LayerNorm ($\epsilon = 10^{-6}$, no affine) \\
QK-Norm & RMSNorm \\
Token dropout $p_{\text{drop}}$ & 0.1 \\
\bottomrule
\end{tabular}
\end{table}

\subsection{3D Shape Tokenizer}
\label{sec:model:3d_tokenizer}
 
A central requirement of our unified architecture is a 3D tokenizer that 
produces discrete tokens capable of faithfully capturing a wide range of 
geometric properties---smooth surfaces, sharp edges, high-frequency 
details---while being natively compatible with autoregressive sequence 
models. We adopt Cube3D~\cite{bhat2025cube}, a shape tokenizer built on the 
3DShape2VecSet~\cite{zhang20233dshape2vecset} framework, which converts continuous 
shape representations into discrete tokens suitable for mixed-modal 
foundation models~\cite{chameleon}.
 
\paragraph{Encoder-Decoder Architecture}
The tokenizer follows an encoder-decoder design. The encoder takes as input a 
point cloud sampled from the mesh surface and embeds it using a phase-modulated 
positional encoding, which improves the Perceiver-based transformer's ability 
to disambiguate spatially distinct points in cross-attention layers. A 
Perceiver-based transformer~\cite{perceiver} compresses the point cloud into 
a set of $L_m = 1{,}024$ continuous latent vectors, which are subsequently 
discretized through optimal-transport vector 
quantization~\cite{bhat2025cube} to produce discrete 3D shape tokens. 
The decoder reconstructs an implicit occupancy field from these discrete tokens, 
from which a triangle mesh can be extracted via Marching Cubes.

\begin{figure}[t]
\centering
\includegraphics[width=\columnwidth]{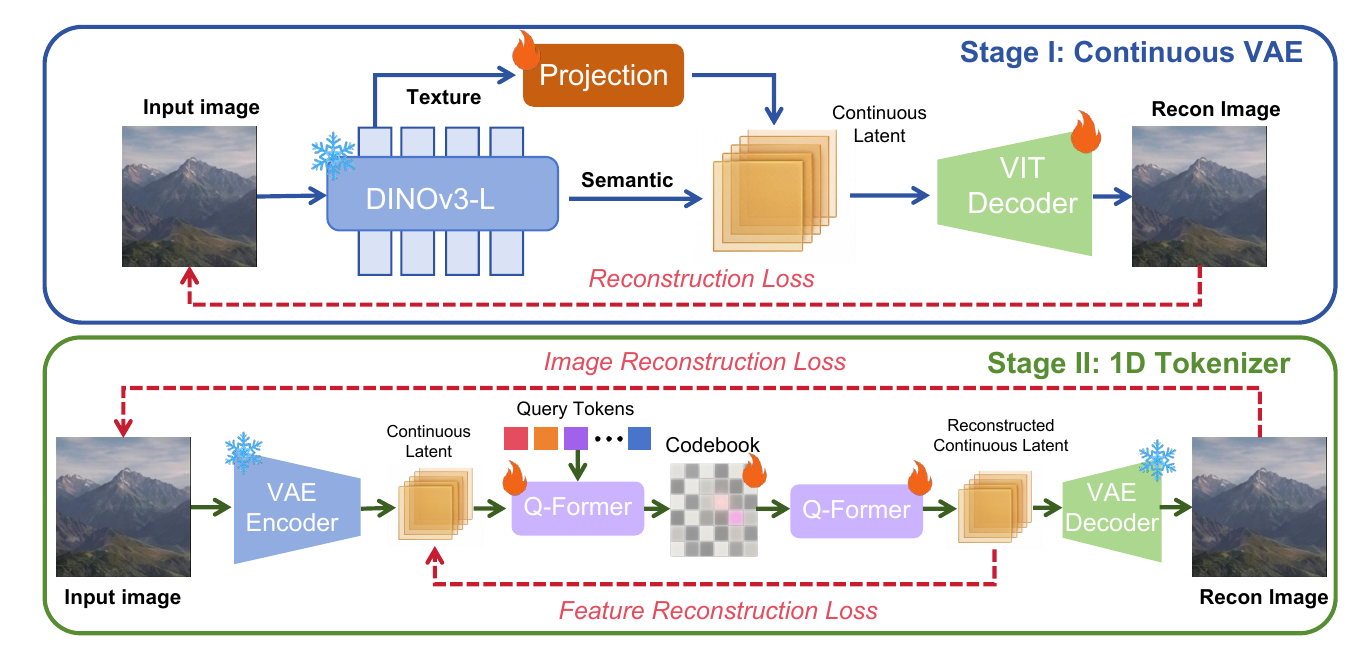}
\caption{Two-stage image tokenizer training strategy. Stage~1 trains a 
continuous VAE (DINO-Tok) to learn high-fidelity visual representations. 
Stage~2 freezes the VAE and trains a 1D Q-Former to reconstruct the 
continuous features, reducing vector quantization to a compact 1D token 
extraction task.}
\label{fig:tokenizer_framework}
\end{figure}

\subsection{Image Tokenizer}
\label{sec:pretraining:image_tokenizer}
Image tokenizers~\cite{vqvae,shi2025scalable-ibq} play a crucial role in compressing visual data into compact discrete latent tokens, facilitating scalable and efficient training of autoregressive models for visual generation via next-token prediction. We build our image tokenizer upon GiGaTok~\cite{gigatok}, whose 1D token representation is more compatible with our 3D shape tokenizer, while also better handling the inherent redundancy in images~\cite{titok}. Recent studies~\cite{simvq} trace the sub-optimal performance of discrete tokenizers back to a core limitation: the disjoint optimization of the codebook. This disjointness inherently degrades reconstruction fidelity compared to continuous VAEs~\cite{rombach2022high, vavae}, ultimately acting as a critical bottleneck that hinders pure autoregressive models from matching the state-of-the-art generation quality of diffusion models or hybrid approaches. To alleviate this issue, we propose a two-stage training strategy (Figure~\ref{fig:tokenizer_framework}): 
First, we train a continuous VAE to learn semantically rich visual representations and high-fidelity reconstruction. Then we insert the 1D GiGaTok~\cite{gigatok} into the pretrained VAE and train it only to reconstruct the continuous features, simplifying the vector quantization to a pure 1D compact token extraction task. This design yields a high-performance tokenizer that is scalable and equipped with rich semantic representations. 

\paragraph{Architectural Details}
Inspired by recent successful Representation Autoencoder\cite{rae,du2025vqrae,shi2025svg}, we retrain DINO-Tok~\cite{jia2025dino} with Dinov3-L~\cite{simeoni2025dinov3} encoder and an 832-dimensional latent space as aforementioned VAE on ImageNet~\cite{deng2009imagenet} with $256 \times 256$ and $512 \times 512$ resolution. In the second stage, we freeze the DINO-Tok VAE and only train the 1D Q-Former on a 60M-image dataset and $512 \times 512$ resolution with 1024 query tokens and a 64K-entry codebook of dimension 64. To the best of our knowledge, our tokenizer is the first 1D visual tokenizer trained at this scale.

\subsection{Model Pre-training (PT)}
\label{sec:model:pretraining}

The central challenge motivating our pre-training design is the acute 
scarcity of 3D data relative to 2D: our corpus contains 63.7\,M 
text--image pairs but only 9\,M text--3D pairs. As argued in 
Section~\ref{sec:intro}, 2D observations implicitly encode rich 3D 
structural information---shape, appearance, and spatial 
relationships---suggesting that large-scale 2D supervision can serve as 
a powerful geometric prior for 3D learning. However, na\"ively mixing appearance-focused and geometry-focused objectives risks task interference rather than beneficial transfer.

Our pre-training strategy addresses this tension through the \emph{X-to-X} 
paradigm: all cross-modal generation tasks---text-to-image, text-to-3D, 
image-to-3D, and 3D-to-image---are unified as autoregressive next-token 
prediction over a shared discrete vocabulary. By training a single model 
to traverse multiple generation paths across the 
``semantic--visual--geometric'' cycle, the framework enforces cross-modal 
generative consistency as an implicit structural constraint. Concretely, the large-scale Text$\to$Image subset strengthens the shared text 
conditioning pathway that Text$\to$3D relies on despite its limited data, 
while the 3D$\to$Image task---as the inverse of Image$\to$3D---forces the 
model to maintain explicit 3D-to-2D geometric correspondences, which in 
turn benefits forward Image$\to$3D reconstruction.

\paragraph{Task Formulation}
During pre-training, we train on four core generation tasks
(Table~\ref{tab:xtox_tasks}). Each task takes one or two source
modalities as conditioning and autoregressively generates tokens in
the target modality. Text, when present, is injected via the
cross-attention pathway; non-text source tokens are placed as a
non-causal prefix in the self-attention sequence. The model is trained
to predict only the target tokens that follow the prefix. Note that
text conditioning is randomly dropped during training to enable
Classifier-Free Guidance (discussed in Training Objective below),
so the model must also learn to generate from the non-text prefix alone.

\begin{table}[t]
\centering
\caption{Pre-training task configurations. Text conditioning is
optionally dropped for Classifier-Free Guidance. The loss is
computed only over the target column.}
\label{tab:xtox_tasks}
\begin{tabular}{@{}llll@{}}
\toprule
\textbf{Task} & \textbf{Cond.\ Stream} & \textbf{Prefix}
& \textbf{Target} \\
\midrule
Text $\to$ Image            & Caption & ---          & Image tokens \\
Text $\to$ 3D               & Caption & ---          & 3D shape tokens \\
(Text +) Image $\to$ 3D     & (Caption) & Image tokens & 3D shape tokens \\
(Text +) 3D $\to$ Image     & (Caption) & 3D shape tokens   & Image tokens \\
\bottomrule
\end{tabular}
\end{table}

\paragraph{Training Objective}
All four tasks share a single autoregressive cross-entropy loss over
the target token sequence:
\begin{equation}
    \mathcal{L} = -\sum_{j=1}^{L_{\text{tgt}}}
    \log\, p_\theta\!\bigl(x_j \mid x_{<j},\, \mathbf{c}\bigr)
    \label{eq:ar_loss}
\end{equation}
where $x_j$ is the $j$-th target token, $x_{<j}$ denotes preceding
tokens including any source-modality prefix, and $\mathbf{c}$ is the
text conditioning (empty when dropped). A uniform loss weight is applied across all tasks. To enable Classifier-Free Guidance~\cite{ho2022classifier} at inference time, we randomly drop the text conditioning with probability
$p_{\text{drop}} = 0.1$ during training~\cite{sun2024autoregressive}, replacing it with
a learned null embedding. For the two prefix-conditioned tasks
(Image$\to$3D and 3D$\to$Image), this means the model is exposed to
three conditioning regimes: text + prefix, prefix only, and unconditional,
allowing flexible guidance at inference.

\paragraph{Data Mixture}
Pre-training draws from three data pools of varying scale: Text--Image
pairs (63.7\,M), Image--3D pairs (120\,M), and Text--3D pairs (9\,M). Each
pool serves one or more tasks: Text--Image supports Text$\to$Image;
Image--3D supports both Image$\to$3D and 3D$\to$Image (with the task
direction randomly selected per sample); and Text--3D supports
Text$\to$3D. Na\"ively sampling in proportion to dataset size would
severely under-represent the smallest yet critical Text--3D subset.
We adopt a temperature-based weighted
sampling strategy. The probability of drawing from subset $i$ is:
\begin{equation}
    P_i = \frac{(\mathrm{Size}_i)^{T} \times \mathrm{Priority}_i}
    {\sum_{j=1}^{n}\bigl((\mathrm{Size}_j)^{T}
    \times \mathrm{Priority}_j\bigr)}
    \label{eq:temp_sampling}
\end{equation}
where $\mathrm{Size}_i$ is the number of samples in pool $i$,
$\mathrm{Priority}_i$ is a manually assigned importance weight
reflecting task value, and $T \in (0,1]$ is a temperature controlling
the degree of rebalancing---lower $T$ flattens the size distribution,
giving smaller but high-priority pools greater representation.
Table~\ref{tab:data_mixture} lists the configuration.

\begin{table}[t]
\centering
\caption{Pre-training data mixture. Each pool may serve multiple tasks
as described in the text.}
\label{tab:data_mixture}
\begin{tabular}{@{}lcccc@{}}
\toprule
\textbf{Data Pool} & \textbf{Size} & \textbf{Tasks Served}
& \textbf{Priority} & \textbf{Effective $P_i$} \\
\midrule
Text--Image   & 63.7\,M & Text$\to$Image
& 1.0 & 0.30 \\
Image--3D     & 120\,M & Image$\to$3D, 3D$\to$Image
& 1.5 & 0.53 \\
Text--3D      & 9\,M  & Text$\to$3D
& 3.0 & 0.17 \\
\bottomrule
\end{tabular}
\end{table}

\paragraph{Training Schedule}
Pre-training proceeds in two stages
(Table~\ref{tab:training_recipe}). Stage~1 trains at $256 \times 256$
resolution for 400{,}000 steps with a peak learning rate of 5e-4,
allowing the model to learn cross-modal alignment across all four tasks.
Stage~2 increases the image resolution to $512 \times 512$ and continues
for 250{,}000 steps with a reduced learning rate of 1e-4, refining
visual fidelity while preserving the learned cross-modal representations.
Both stages use AdamW~\cite{loshchilov2017decoupled} ($\beta_1 = 0.9$,
$\beta_2 = 0.999$, weight decay $= 0.1$) with a cosine learning rate
schedule. Gradient norms are clipped at $5.0$. The effective batch size
is 2{,}048 (Stage~1) and 1{,}024 (Stage~2) across 256$\times$H100 GPUs.
In total, the model sees approximately 1.16\,T tokens during pre-training.

\begin{table}[t]
\centering
\caption{Training recipe across all stages.}
\label{tab:training_recipe}
\begin{tabular}{@{}lcccc@{}}
\toprule
& \multicolumn{2}{c}{\textbf{Pre-training}} & \multicolumn{2}{c}{\textbf{Post-training}} \\
\cmidrule(lr){2-3} \cmidrule(lr){4-5}
\textbf{Hyperparameter} & \textbf{Stage 1} & \textbf{Stage 2} & \textbf{CT} & \textbf{SFT} \\
\midrule
Peak learning rate   & 5e-4 & 1e-4 & 1e-5 & 1e-6 \\
LR scheduler         & \multicolumn{4}{c}{Cosine} \\
Optimizer            & \multicolumn{4}{c}{AdamW ($\beta_1\!=\!0.9$, $\beta_2\!=\!0.999$)} \\
Weight decay         & \multicolumn{4}{c}{0.1} \\
Gradient norm clip   & \multicolumn{4}{c}{5.0} \\
Warm-up steps        & 5{,}000 & 1{,}000 & 1{,}000 & 2{,}000 \\
Training steps       & 400{,}000 & 250{,}000 & 120{,}000 & 115{,}000 \\
Effective batch size & 2{,}048 & 1{,}024 & 1{,}024 & 256 \\
Image resolution     & $256^2$ & $512^2$ & $512^2$ & $512^2$ \\
GPUs                 & \multicolumn{4}{c}{256$\times$H100} \\
Tokens seen          & ${\sim}0.76$\,T & ${\sim}0.40$\,T & ${\sim}0.20$\,T & ${\sim}0.15$\,T \\
\bottomrule
\end{tabular}
\end{table}

\subsection{Continued Training (CT)}
\label{sec:ct}

Pre-training establishes cross-modal alignment between text, image, and  3D representations, yet the model lacks an explicit notion of 
\emph{viewpoint}: it can generate images and 3D shapes, but cannot 
reason about how the same object appears from different camera poses. 
This is a critical gap---multi-view consistency is the bridge that 
connects 2D appearance to 3D geometry, and without it, the 
``semantic--visual--geometric'' cycle introduced in 
Section~\ref{sec:intro} remains incomplete. Continued training (CT) 
closes this gap by introducing \emph{view tokens} 
(Figure~\ref{fig:ct}): a set of $N$ learnable embeddings, each bound 
to a fixed canonical viewpoint (front, back, left, right, top, bottom). 
By prepending a view token to the target image sequence, the model 
learns to associate each embedding with a specific camera pose, enabling 
viewpoint-controllable generation that is geometrically grounded in the 
shared 3D representation learned during pre-training.

\paragraph{Task Formulation}
CT retains all four pre-training tasks and introduces three new 
view-conditioned tasks (Figure~\ref{fig:ct}): Image $\to$ Posed Image, 
3D $\to$ Posed Image, and Text $\to$ Posed Image. The new tasks share 
a common mechanism: a learnable view token $\mathbf{v}$ encoding the 
target camera extrinsics are prepended to the output image tokens in the 
generation stream. When the source is also an image, it is accompanied 
by its own view token $\mathbf{v}_{\text{src}}$ in the prefix, 
providing the model with explicit source--target viewpoint 
correspondence and enabling novel view synthesis. This design yields 
progressively richer 3D-aware capabilities---from single-image novel 
view synthesis (Image $\to$ Posed Image) to explicit 3D-controllable 
rendering (3D $\to$ Posed Image) to text-driven viewpoint-specific 
generation (Text $\to$ Posed Image).

\begin{figure*}[t]
\centering
\includegraphics[width=\textwidth]{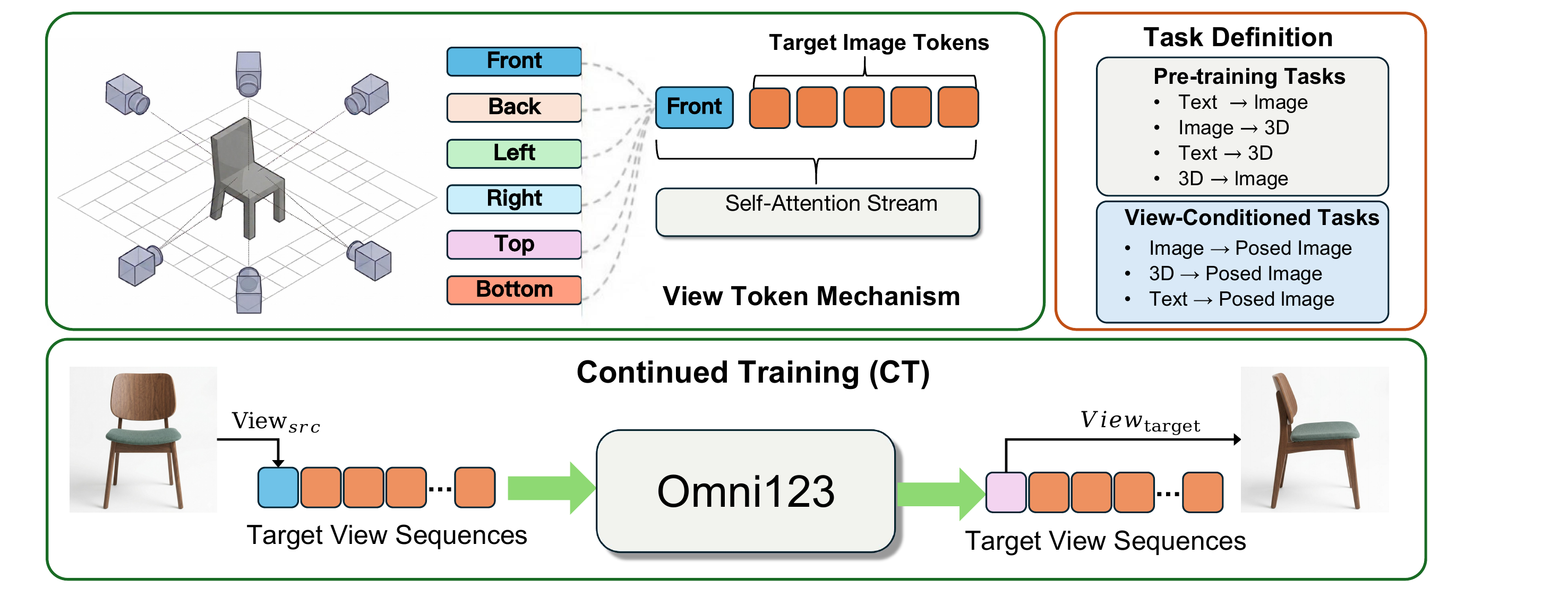}
\caption{Continued training introduces view-conditioned generation via learnable 
view tokens. \textbf{Top left:} $N{=}6$ view tokens corresponding to canonical 
viewpoints (front, back, left, right, top, bottom) are mapped from 3D camera extrinsics. \textbf{Top right:} A view token $\mathbf{v}$ is prepended to the 
target image tokens and fed into the self-attention stream, enabling 
viewpoint-controllable generation. \textbf{Bottom:} In the Posed Image 
$\to$ Posed Image task, the source image is accompanied by its own view token 
$\mathbf{v}_{\text{src}}$, providing explicit source--target viewpoint 
correspondence for novel view synthesis.}
\label{fig:ct}
\end{figure*}

\paragraph{Data}
The view-conditioned tasks are constructed from the existing Image--3D
pairs. For each 3D asset, we render images at $N=6$ fixed canonical
viewpoints (front, back, left, right, top, bottom), yielding posed
image--3D tuples. During training, a task and a target viewpoint are
randomly sampled per example. The pre-training data pools remain active
to prevent catastrophic forgetting of the base generation capabilities.

\paragraph{Training Setup}
CT uses the same training objective (Eq.~\ref{eq:ar_loss}) and optimizer
configuration as pre-training. Only the $N$ view token embeddings are
newly initialized; all other parameters are warm-started from the
pre-trained checkpoint. The learning rate is reduced to $1 \times 10^{-5}$ with 1{,}000
warm-up steps. The model trains for 120{,}000 steps with an effective batch
size of 1{,}024. The temperature-based data mixing strategy
(Eq.~\ref{eq:temp_sampling}) is extended to include the new
view-conditioned pool, with priority weights adjusted to emphasize
3D-aware generation while retaining base capabilities.

\subsection{Supervised Fine-Tuning (SFT)}
\label{sec:posttraining:sft}

Pre-training and Continued Training equip the model with cross-modal alignment and 
viewpoint awareness, but each task is still trained on pairwise data 
independently: there is no guarantee that a 3D mesh produced by 
Image$\to$3D renders back to the conditioning image, or that an image 
generated by Text$\to$Image yields a plausible 3D reconstruction.
The SFT stage closes this loop by introducing \emph{interleaved 
multi-modal sequences} that chain image generation, 3D reconstruction, 
and novel view rendering into a single autoregressive sequence. Because 
each subsequent modality is conditioned on all previously generated 
tokens, the training signal naturally enforces cross-modal consistency: 
a 3D mesh must be faithful to the images that precede it, and novel 
views rendered from that mesh must be coherent with the original 
generation.

\paragraph{Task Formulation}
We define five SFT tasks (Table~\ref{tab:sft_tasks}) covering all 
practical generation pipelines. Each task is formulated as a single 
autoregressive sequence over interleaved modality tokens. Text 
conditioning remains in the conditioning stream with the same CFG 
dropout ($p_{\text{drop}}=0.1$). View tokens $\mathbf{v}_k$ specify 
camera poses for posed image generation.

\begin{table}[t]
\centering
\caption{SFT task configurations. All tasks use text conditioning 
via the conditioning stream. $\mathbf{v}_k$ denotes view tokens. 
Img$_{\times N}$ denotes $N$ posed images generated sequentially.}
\label{tab:sft_tasks}
\setlength{\tabcolsep}{4pt}
\begin{tabular}{@{}p{0.35\linewidth}p{0.6\linewidth}@{}}
\toprule
\centering\textbf{Task} & \centering\textbf{Sequence}\arraybackslash \\
\midrule
\centering Text $\to$ Img $\to$ 3D
& \centering $[\text{Img}_1,\, \text{3D}]$ \arraybackslash \\[4pt]
\centering Text $\to$ 3D $\to$ Img
& \centering $[\text{3D},\, \mathbf{v}_1,\, \text{Img}_1]$ \arraybackslash \\[4pt]
\centering Img$_{\times N}$ $\to$ 3D
& \centering $[\mathbf{v}_1,\, \text{Img}_1,\, \ldots,\,
   \mathbf{v}_N,\, \text{Img}_N,\, \text{3D}]$ \arraybackslash \\[4pt]
\centering 3D $\to$ Img$_{\times N}$
& \centering $[\text{3D},\, \mathbf{v}_1,\, \text{Img}_1,\, \ldots,\,
   \mathbf{v}_N,\, \text{Img}_N]$ \arraybackslash \\[4pt]
\centering Text $\to$ Img$_{\times N}$ $\to$ 3D $\to$ Img$_{\times M}$
& \centering $[\mathbf{v}_1,\, \text{Img}_1,\, \ldots,\,
   \mathbf{v}_N,\, \text{Img}_N,\, \text{3D},\,
   \mathbf{v}_{N\!+\!1},\, \text{Img}_{N\!+\!1},\, \ldots]$ \arraybackslash \\
\bottomrule
\end{tabular}
\end{table}

\section{Align to Instruction-Based 3D Editing}
\label{sec:editing}
Earlier sections have demonstrated that \methodname{}'s unified autoregressive architecture excels not only at 3D generation but also at cross-modal understanding, owing to its inherited text-conditioned capabilities and the rich 3D--2D correspondences learned during pre-training and continued training. Consequently, we can adapt the model into an instruction-based 3D editing system, further revealing its potential to benefit users in practical content-creation workflows.

\paragraph{Task Formulation}
We formulate instruction-based 3D editing as a conditional generation task as follows. Let $\mathcal{M}_{\text{src}}$ denote the source mesh tokenized into $L_m = 1{,}024$ discrete tokens $\mathbf{s}$ via Cube3D, and let $c_{\text{edit}}$ be a natural-language edit instruction (e.g., ``add wings to the car'', ``change the material to gold''). Following the prefix-conditioned paradigm established in pre-training (Section~\ref{sec:model:pretraining}), the source 3D shape tokens are placed as a non-causal prefix in the self-attention stream, while the edit instruction is injected via cross-attention. The model autoregressively generates the target token sequence $\mathbf{s}' = (s'_1, \dots, s'_{L_m})$, which is decoded into the edited mesh $\mathcal{M}_{\text{tgt}}$ via the Cube3D decoder and Marching Cubes. Text conditioning is randomly dropped with probability $p_{\text{drop}} = 0.1$ during training to enable Classifier-Free Guidance at inference, so the model is exposed to three conditioning regimes: edit instruction + source prefix, source prefix only, and unconditional, mirroring the CFG setup of the prefix-conditioned pre-training tasks.

\paragraph{Data}
A key bottleneck for instruction-based 3D editing has been the lack of large-scale, high-quality paired data. Unlike 2D image editing, where paired datasets can be constructed at scale using foundation models~\cite{instructpix2pix}, 3D editing data must additionally ensure cross-view consistency, structural fidelity in unedited regions, and semantic coherence of the edited parts. We adopt 3DEditVerse~\cite{xia2025towards}, the largest paired 3D editing benchmark to date, comprising 116{,}309 high-quality training pairs and 1{,}500 curated test pairs. Each pair consists of a source mesh, an edited mesh, and a natural-language edit instruction describing the modification. 3DEditVerse is constructed through two complementary automated pipelines: \emph{(i)} a \emph{pose-driven geometric} pipeline that leverages publicly available 3D characters and animation sequences, exploiting the fact that different poses of the same character naturally form valid ``before--after'' edit pairs for articulation and structural variations; and \emph{(ii)} a \emph{foundation-model-guided appearance} pipeline that orchestrates multiple foundation models to generate edited assets with modified textures, colors, or stylistic attributes while preserving the underlying geometry. Both pipelines ensure edit locality, multi-view consistency, and semantic alignment between the instruction and the applied modification. Each training pair is tokenized into source tokens $\mathbf{s}$, target tokens $\mathbf{s}'$, and text embeddings via the same encoders used throughout \methodname{}.
 
\paragraph{Training Objective}
The editing task shares the same autoregressive cross-entropy loss as all earlier training stages (Eq.~\ref{eq:ar_loss}), applied over the target token sequence:
\begin{equation}
    \mathcal{L}_{\text{edit}} =
    -\sum_{k=1}^{L_m}
    \log\, p_\theta\!\bigl(s'_k \mid s'_{<k},\,
    \mathbf{s},\, c_{\text{edit}}\bigr)
    \label{eq:edit_loss}
\end{equation}
where $s'_k$ is the $k$-th target 3D token, $s'_{<k}$ denotes all preceding target tokens, $\mathbf{s}$ is the source prefix, and $c_{\text{edit}}$ is the text conditioning (replaced with a learned null embedding when dropped). The loss is computed only over the target tokens; the source prefix is excluded from the loss computation. A uniform loss weight is applied across all target positions.
 
\paragraph{Training Setup}
All parameters are initialized from the SFT checkpoint (Section~\ref{sec:posttraining:sft}). We fine-tune the full model using AdamW~\cite{loshchilov2017decoupled} ($\beta_1 = 0.9$, $\beta_2 = 0.999$, weight decay $= 0.1$) with a peak learning rate of $1 \times 10^{-5}$ and a cosine schedule with 1{,}000 warm-up steps. Training proceeds for 10{,}000 steps with an effective batch size of 128 across 32$\times$H100 GPUs. Gradient norms are clipped at $5.0$, consistent with the pre-training configuration.

\begin{figure*}[t]
\centering
\includegraphics[width=\textwidth,height=1.4\textwidth]{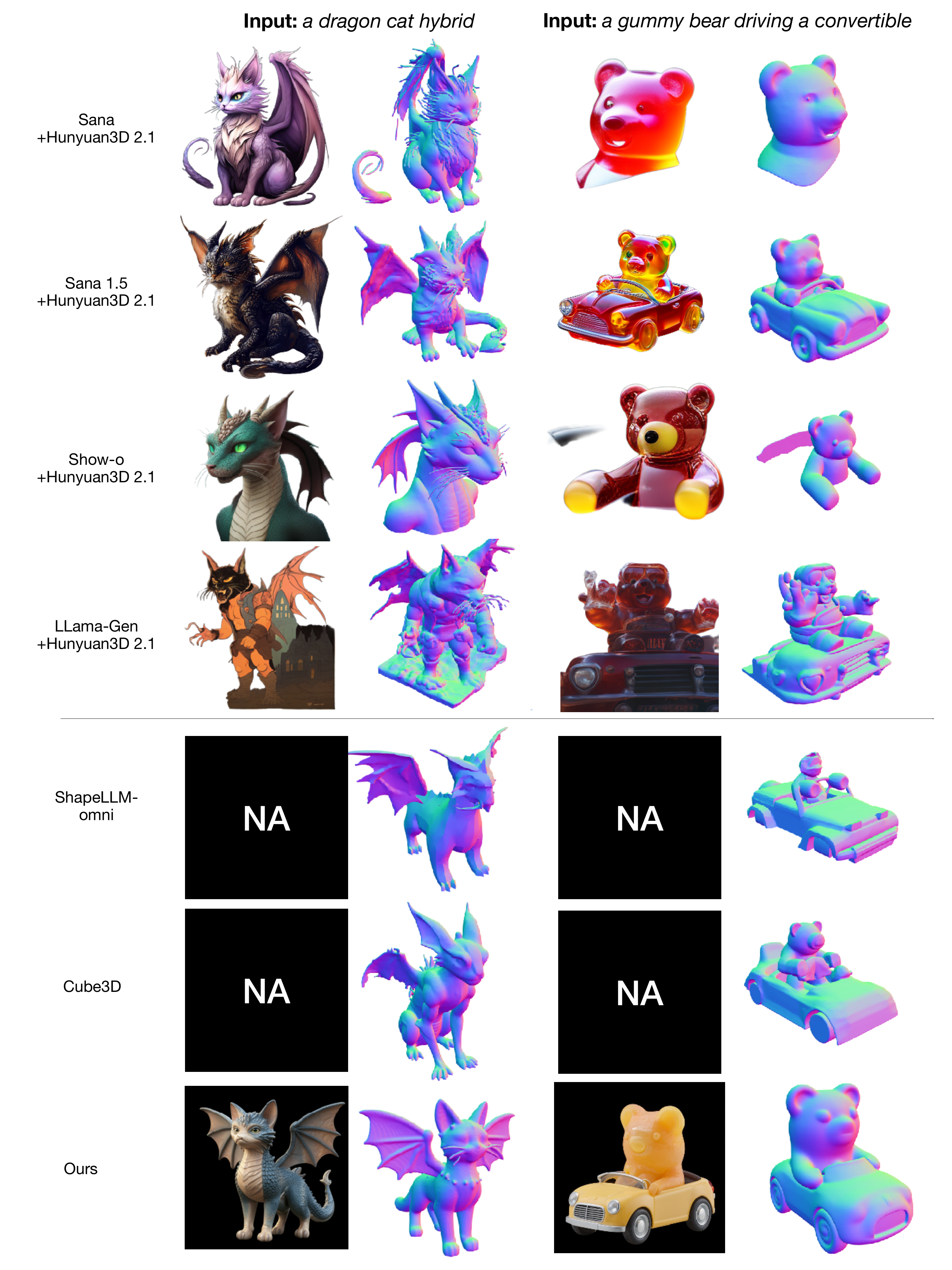}
\caption{Qualitative comparison of text-to-3D shape generation across 
two compositional prompts. \emph{Rows 1--4}: Two-stage 
Text$\to$Image$\to$3D methods, each showing the generated image (left) 
and reconstructed mesh as rendered image and normal map (right). 
\emph{Rows 5--6}: Native text-to-3D baselines, which generate only 
meshes without intermediate images (marked ``NA''). \emph{Row 7}: 
\methodname{} (Ours), which natively generates both a rendered image 
and a 3D mesh within a single autoregressive sequence.}
\label{fig:sota_comparison}
\end{figure*}
\section{Experiments}
\label{sec:experiments}

We evaluate \methodname{} on text-to-3D shape generation (Section~\ref{sec:eval:text_to_3d}), instruction-based 3D editing (Section~\ref{sec:eval:editing}), and image tokenizer reconstruction quality (Section~\ref{sec:eval:tokenizer}). Unless otherwise stated, all \methodname{} results use greedy decoding with Classifier-Free Guidance scale $\omega = 5.0$.
 
\subsection{Text-to-3D Shape Generation}
\label{sec:eval:text_to_3d}
 
As a cornerstone of 3D content generation, structurally consistent and high-quality shape generation provides the essential geometric scaffolding required for subsequent pipelines, including texturing, rendering, and physics simulation. Here, we evaluate the text-to-shape generation performance of \methodname{}, focusing on geometric fidelity and semantic alignment with input prompts.
 
\paragraph{Metrics}
To evaluate semantic alignment, we uniformly sample $8{,}192$ surface points from each generated mesh and compute point-text similarities using two established multi-modal models: ULIP~\cite{ulip}(reporting ULIP-T) and Uni3d~\cite{uni3d}(reporting Uni3D-T). We evaluate on 100 diverse text prompts encompassing various object categories and styles, compositional descriptions, and fine-grained attributes.
 
\paragraph{Baselines}
To rigorously isolate the advantages of our \textbf{interleaved X-to-X 
training paradigm}, we benchmark against two dominant methodologies.
\textbf{\textit{(i) Cascaded text-to-image-to-3D pipelines:}} We pair 
leading image generators of \textit{comparable parameter 
scales}---Sana-1.5~\cite{sana1.5} (4.8B), Sana~\cite{sana} (1.6B), 
Show-o~\cite{xie2024show} (1.3B), and 
LlamaGen~\cite{sun2024autoregressive} (0.775B)---with 
Hunyuan3D-2.1~\cite{hunyuan3d-2.1}. These two-stage baselines represent 
indirect workflows that frequently compromise geometric consistency.
\textbf{\textit{(ii) Native text-to-3D models:}} 
ShapeLLM-Omni~\cite{ye2025shapellm}, a 7B autoregressive model that 
extends a pre-trained VLM (Qwen2.5-VL~\cite{qwen2.5-vl}) with 3D 
generation capabilities, and Cube3D~\cite{bhat2025cube}, a 1.6B 
autoregressive model trained natively for text-to-3D generation. Both 
bypass the two-stage handoff but are constrained by 3D data scarcity 
and limited cross-modal transfer.

\paragraph{Quantitative Comparison}
Table~\ref{tab:text_to_3d} demonstrates that two-stage pipelines consistently underperform native methods in \textbf{semantic-geometric alignment}, primarily bottlenecked simultaneously by the inherent prior gap between the image generator and the 3D lifter and inevitable error accumulation across the cascaded pipeline. Conversely, VLM-based native models avoid this handoff but are constrained by 3D data scarcity, ineffective knowledge transfer, and sub-optimal modality fusion. By leveraging abundant 2D observations as a unified geometric prior, both \methodname{} variants surpass these baselines by a clear margin. Notably, our \textbf{2B model} achieves superior alignment compared to the 7B ShapeLLM-Omni~\cite{ye2025shapellm}, demonstrating that interleaved cross-modal training is a fundamentally more \emph{parameter-efficient}, scalable route than blindly scaling isolated unimodal generators upon a pretrained VLM backbone. 

\paragraph{Qualitative Comparison}
Figure~\ref{fig:sota_comparison} visually supports our quantitative 
findings on complex compositional prompts. Two-stage pipelines produce 
compelling intermediate images, but their image-to-3D handoff frequently 
collapses into artifact-ridden meshes---geometric flattening, disjointed 
limbs, and loss of fine-grained structure are common failure modes. 
Native 3D baselines bypass this fragile lifting step, yet their reliance on scarce 3D data limits compositional generalization: complex prompts 
often result in entangled features and semantically unfaithful geometry. 
\methodname{} avoids both failure modes by generating a high-fidelity 2D image and a structurally sound mesh within a single autoregressive 
sequence, successfully articulating fine-grained anatomical features and precise multi-object spatial relationships.

\begin{table}[t]
\centering
\caption{Quantitative comparison of text-to-3D shape generation.
We evaluate shape--text alignment using ULIP-T~\cite{ulip}
and Uni3D-T~\cite{uni3d}.
The two-stage methods pair various image generators with
Hunyuan3D\,2.1~\cite{hunyuan3d-2.1} as the image-to-3D backbone.
Best overall is \textbf{bolded}, second-best is \underline{underlined}.}
\label{tab:text_to_3d}
\setlength{\tabcolsep}{6pt}
\begin{tabular}{@{}lrcc@{}}
\toprule
\textbf{Method}
& \textbf{Params}
& \textbf{ULIP-T\,$\uparrow$}
& \textbf{Uni3D-T\,$\uparrow$} \\
\midrule
\multicolumn{4}{@{}l}{\emph{Two-stage: Text\,$\to$\,Image\,$\to$\,3D
  \;\;(3D backbone: Hunyuan3D\,2.1)}} \\[3pt]
\quad Sana-1.5~\cite{sana1.5}
  & 4.8B  & 0.1516 & 0.1672 \\
\quad Sana~\cite{sana}
  & 1.6B  & 0.1448 & 0.1589 \\
\quad Show-o~\cite{xie2024show}
  & 1.3B  & 0.1447 & 0.1581 \\
\quad LlamaGen~\cite{sun2024autoregressive}
  & 0.775B  & 0.0946 & 0.1067 \\
\midrule
\multicolumn{4}{@{}l}{\emph{Native: Text\,$\to$\,3D}} \\[3pt]
\quad ShapeLLM-Omni~\cite{ye2025shapellm}
  & 7B    & 0.1649 & 0.2666 \\
\quad Cube3D~\cite{bhat2025cube}
  & 1.6B  & 0.1726 & 0.2768 \\
\quad \textbf{Ours}
  & 2.2B   & \textbf{0.1832} & \underline{0.2855} \\
\bottomrule
\end{tabular}
\end{table}

\begin{table}[t]
\centering
\caption{Quantitative comparison of image tokenizers on ImageNet-1K~\cite{deng2009imagenet} $512\times512$ test set.}
\label{tab:recon_quality_all}
\setlength{\tabcolsep}{10pt}
\renewcommand{\arraystretch}{1.15}
\begin{tabular}{lccc}
\toprule
Method & FID $\downarrow$ & PSNR $\uparrow$ & SSIM $\uparrow$ \\
\midrule
\multicolumn{4}{l}{\textbf{\textit{Continuous}}} \\
Cosmos-CI$16\times16$~\cite{agarwal2025cosmos} & 0.356 & 28.60 & 0.7826  \\
Flux.1~\cite{flux2024} & \textbf{0.017} & 37.69 & 0.963 \\
SDXL~\cite{podell2023sdxl} & 0.188 & 30.35 & 0.835 \\
SVG-T2I~\cite{shi2025svg} & 1.994 & 20.60 & 0.587 \\
\textbf{Ours VAE} & \textbf{0.017} & \textbf{39.15} & \textbf{0.966} \\
\midrule
\multicolumn{4}{l}{\textbf{\textit{Discrete}}} \\
Cosmos-DI$16\times16$~\cite{agarwal2025cosmos} & 1.215 & 23.76 & 0.645 \\
LlamaGen~\cite{sun2024autoregressive} & 0.839 & 23.96 & 0.664 \\
Show-o~\cite{xie2024show} & 1.183 & 24.20 & 0.680 \\
\textbf{Ours VAE + 1D Tokenizer} & \textbf{0.432} & \textbf{24.61} & \textbf{0.759} \\
\specialrule{1.2pt}{0pt}{0pt}
\end{tabular}
\end{table}

\subsection{Instruction-Based 3D Editing}
\label{sec:eval:editing}
As the most direct and natural interface for 3D creation, language instruction-driven 3D editing is a pivotal capability for broad 3D applications. We evaluate \methodname{} on Edit3D-Bench~\cite{ma2025feedforward}, utilizing its curated source-target mesh pairs for geometric \emph{addition} and \emph{removal} operations. Our goal is to perform semantically faithful and structurally articulated modifications to a source mesh closely following language instructions.

\paragraph{Metrics}
To quantify editing fidelity, we report Chamfer Distance (CD) and F1 score (F1) between the generated and ground-truth edited meshes. Each test sample requires the model to resolve natural-language directives into precise, localized geometric transformations.
 
\paragraph{Baselines}
We compare against four recent 3D editing methods: ShapeLLM-Omni~\cite{ye2025shapellm}, Steer3D~\cite{ma2025feedforward}, 3DEditFormer~\cite{xia2025towards}, and Tailor3D~\cite{qi2024tailor3d}. This selection strategically spans both native autoregressive editors and optimization-based pipelines. While ShapeLLM-Omni features a native text-guided variant, its publicly available weights only support image-based inputs. To ensure maximum fairness and isolate each model’s intrinsic 3D editing and instruction-following capabilities, we intentionally employ the powerful Flux-Kontext~\cite{flux-kontext} to provide high-quality 2D edited image input.
 
\paragraph{Quantitative Comparison}
Table~\ref{tab:edit_results} highlights our superior geometric fidelity. 
\methodname{} achieves the lowest Chamfer Distance (CD) across all tasks, 
indicating significantly tighter structural alignment with the ground-truth 
targets. While Steer3D leads in F1 score, we note that Edit3D-Bench is 
constructed using the same data pipeline as Steer3D's training set, giving 
it a distributional advantage on this benchmark. Despite this, our consistent 
lead in CD demonstrates more precise execution of global geometric 
transformations and topological changes. These results suggest that the 
cross-modal representations learned through interleaved training translate 
effectively to editing tasks, enabling faithful instruction following even 
with relatively limited fine-tuning data (116K pairs).

\begin{table*}[t]
\centering
\caption{Comparison on add and remove editing tasks in Edit3D-Bench~\cite{ma2025feedforward}.}
\label{tab:edit_results}
\setlength{\tabcolsep}{10pt}
\renewcommand{\arraystretch}{1.15}
\begin{tabular}{lcccccc}
\hline
\multirow{2}{*}{Method} & \multicolumn{2}{c}{Add} & \multicolumn{2}{c}{Remove} & \multicolumn{2}{c}{Avg} \\
\cline{2-7}
& CD $\downarrow$ & F1 $\uparrow$ & CD $\downarrow$ & F1 $\uparrow$ & CD $\downarrow$ & F1 $\uparrow$ \\
\hline
\textbf{Ours} & \textbf{0.0736} & 0.1743 & \textbf{0.0632} & 0.2259 & \textbf{0.0684} & 0.2001 \\
ShapeLLM-Omni~\cite{ye2025shapellm} & 0.2546 & 0.0877 & 0.2237 & 0.1166 & 0.2392 & 0.1022 \\
Steer3D~\cite{ma2025feedforward} & 0.1404 & \textbf{0.2414} & 0.0976 & \textbf{0.3044} & 0.1190 & \textbf{0.2729} \\
3DEditFormer~\cite{xia2025towards} & 0.1676 & 0.1955 & 0.1342 & 0.1836 & 0.1509 & 0.1896 \\
Tailor3D~\cite{qi2024tailor3d} & 0.1661 & 0.1217 & 0.1755 & 0.1352 & 0.1708 & 0.1285 \\
\hline
\end{tabular}
\end{table*}

\subsection{Image Tokenizer Evaluation}
\label{sec:eval:tokenizer}

\begin{figure*}[t]
  \centering
  \includegraphics[width=\textwidth]{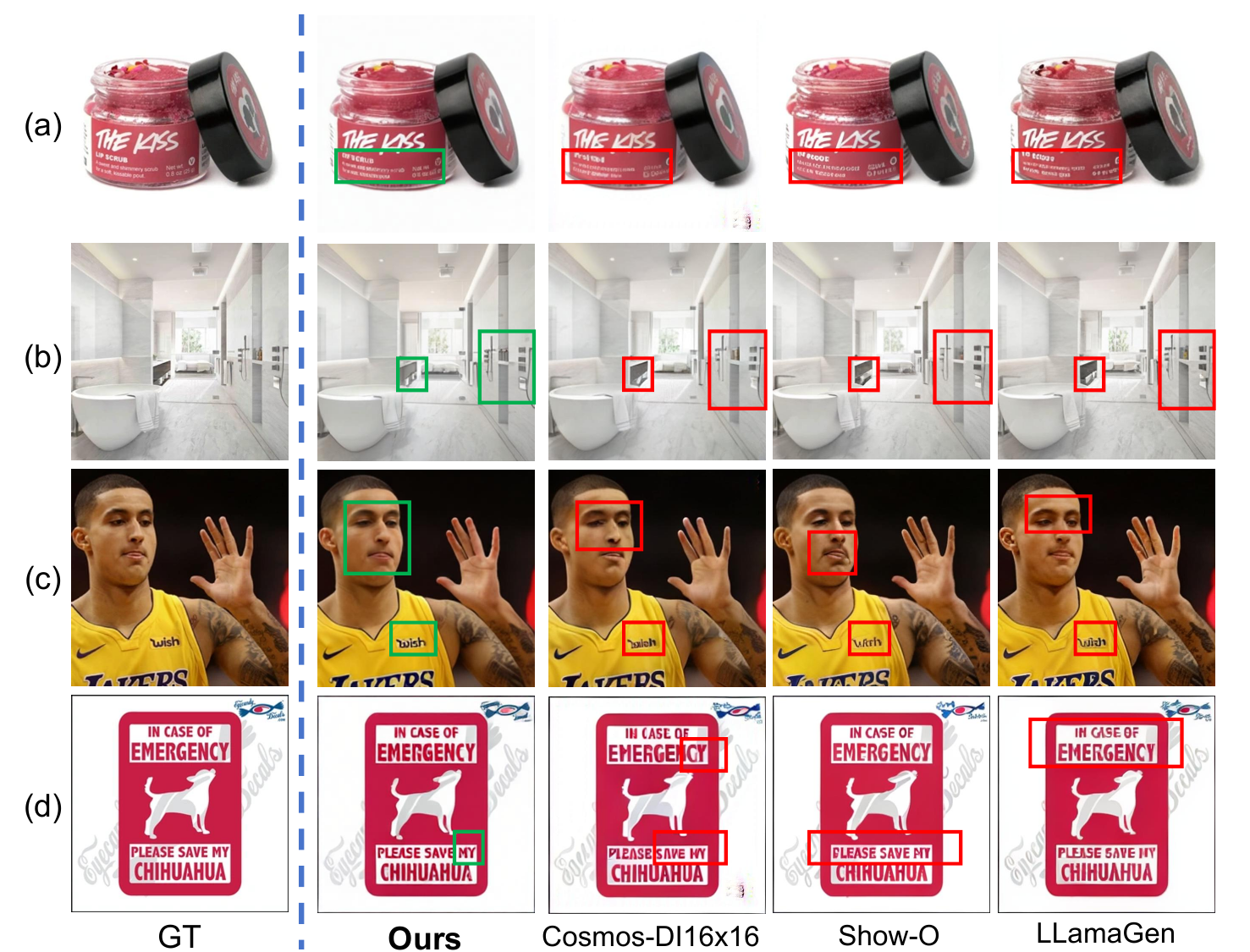}
  \caption{Qualitative comparison of discrete image tokenizers on web images of $512\times512$ resolution.\textbf{Our tokenizer} demonstrates superior reconstruction quality in \textbf{(a)} object-centric images, \textbf{(b)} scene-level details, \textbf{(c)} human details, and \textbf{(d)} text rendering.}
  \label{fig:tokenizer_compare}
\end{figure*}
 
As the representational foundation of our unified framework, a high-fidelity image tokenizer dictates essential structural and textural constraints and mediates the cross-modal information flow. We evaluate our two-stage tokenizer (Section~\ref{sec:pretraining:image_tokenizer}) on the ImageNet-1K~\cite{deng2009imagenet} $512 \times 512$ test set, benchmarking both the continuous VAE (Stage~1) and the full discrete 1D pipeline.

\begin{table}[t]
\centering
\caption{Quantitative comparison of image tokenizers on ImageNet-1K~\cite{deng2009imagenet} $512\times512$ test set.}
\label{tab:recon_quality_all}
\setlength{\tabcolsep}{10pt}
\renewcommand{\arraystretch}{1.15}
\begin{tabular}{lccc}
\toprule
Method & FID $\downarrow$ & PSNR $\uparrow$ & SSIM $\uparrow$ \\
\midrule
\multicolumn{4}{l}{\textbf{\textit{Continuous}}} \\
Cosmos-CI$16\times16$~\cite{agarwal2025cosmos} & 0.356 & 28.60 & 0.7826  \\
Flux.1~\cite{flux2024} & \textbf{0.017} & 37.69 & 0.963 \\
SDXL~\cite{podell2023sdxl} & 0.188 & 30.35 & 0.835 \\
SVG-T2I~\cite{shi2025svg} & 1.994 & 20.60 & 0.587 \\
\textbf{Ours VAE} & \textbf{0.017} & \textbf{39.15} & \textbf{0.966} \\
\midrule
\multicolumn{4}{l}{\textbf{\textit{Discrete}}} \\
Cosmos-DI$16\times16$~\cite{agarwal2025cosmos} & 1.215 & 23.76 & 0.645 \\
LlamaGen~\cite{sun2024autoregressive} & 0.839 & 23.96 & 0.664 \\
Show-o~\cite{xie2024show} & 1.183 & 24.20 & 0.680 \\
\textbf{Ours VAE + 1D Tokenizer} & \textbf{0.432} & \textbf{24.61} & \textbf{0.759} \\
\specialrule{1.2pt}{0pt}{0pt}
\end{tabular}
\end{table}
\paragraph{Metrics}
We evaluate the tokenizer's performance across three hierarchical dimensions: FID to measure the distributional alignment between reconstructed and ground-truth image manifolds, ensuring results are perceptually natural, realistic, and statistically sound; PSNR to quantify pixel-level fidelity; and SSIM to assess structural similarity and geometric integrity. 

\paragraph{Quantitative Comparison}
Table~\ref{tab:recon_quality_all} highlights our superior reconstruction fidelity. In the continuous regime, our VAE surpasses the renowned Flux.1 in fidelity and structural consistency \textbf{using only ImageNet training set}. This underscores the exceptionally powerful prior inherited from the pretrained representation encoder (DINOv3~\cite{simeoni2025dinov3}). Similarly, our discrete pipeline significantly outperforms recent state-of-the-art tokenizers by a large margin across all metrics. These results validate our two-stage strategy as an effective means of minimizing quantization degradation while ensuring a robust representational space.

\paragraph{Qualitative Comparison}
Figure~\ref{fig:tokenizer_compare} illustrates the outstanding reconstruction quality achieved by our image tokenizer. While competing methods often suffer from significant blurriness and semantic distortion, \methodname{} demonstrates a clear superiority in preserving high-frequency details and structural integrity. Notably, our model achieves \textbf{highly faithful text rendering and logo reconstruction}(Rows \textit{a}, \textit{c}, and \textit{d}), which are particularly challenging for discrete tokenizers. Moreover, in \textbf{complex scenes} (Row \textit{b}) and \textbf{human portraits} (Row \textit{c}), our tokenizer preserves \textit{sharp geometric boundaries} and \textit{identity-consistent facial features}. These qualitative results are well aligned with the quantitative comparisons, further confirming that our \textbf{1D image tokenizer} effectively retains the geometric and semantic priors necessary to provide \textbf{high-fidelity guidance} for joint 3D synthesis.

\section{Conclusion}
\label{sec:conclusion}

We presented \methodname{}, a unified multimodal generative model that 
addresses 3D data scarcity by unifying text-to-2D and text-to-3D 
generation within a single autoregressive framework. Our three-stage 
training recipe---cross-modal pre-training over heterogeneous pairs, 
continued training with learnable view tokens for viewpoint awareness, 
and Text--Image--3D interleaved fine-tuning that closes the 
``semantic--visual--geometric'' loop---enables abundant 2D observations 
to serve as an implicit structural constraint for 3D learning. 
Experiments on text-guided 3D generation and instruction-based 3D 
editing to validate that this unified approach significantly improves 
geometric fidelity and semantic alignment, offering a scalable pathway 
toward multimodal 3D world models. Current limitations include the 
fixed mesh resolution of the 3D tokenizer and reliance on canonical 
viewpoints; future work will explore adaptive-resolution tokenization, 
scene-level generation, and material and physics modeling.

\paragraph{Acknowledgments} We thank Xiaohe Ma, Bing Yu, Jianfeng Xiang for insightful discussions and valuable suggestions. 
We thank the Meshy AI team for their contributions to infrastructure,
data curation, and evaluation.


\clearpage
\begin{figure}[p]
  \centering
  \includegraphics[width=\textwidth,height=0.9\textheight,keepaspectratio]{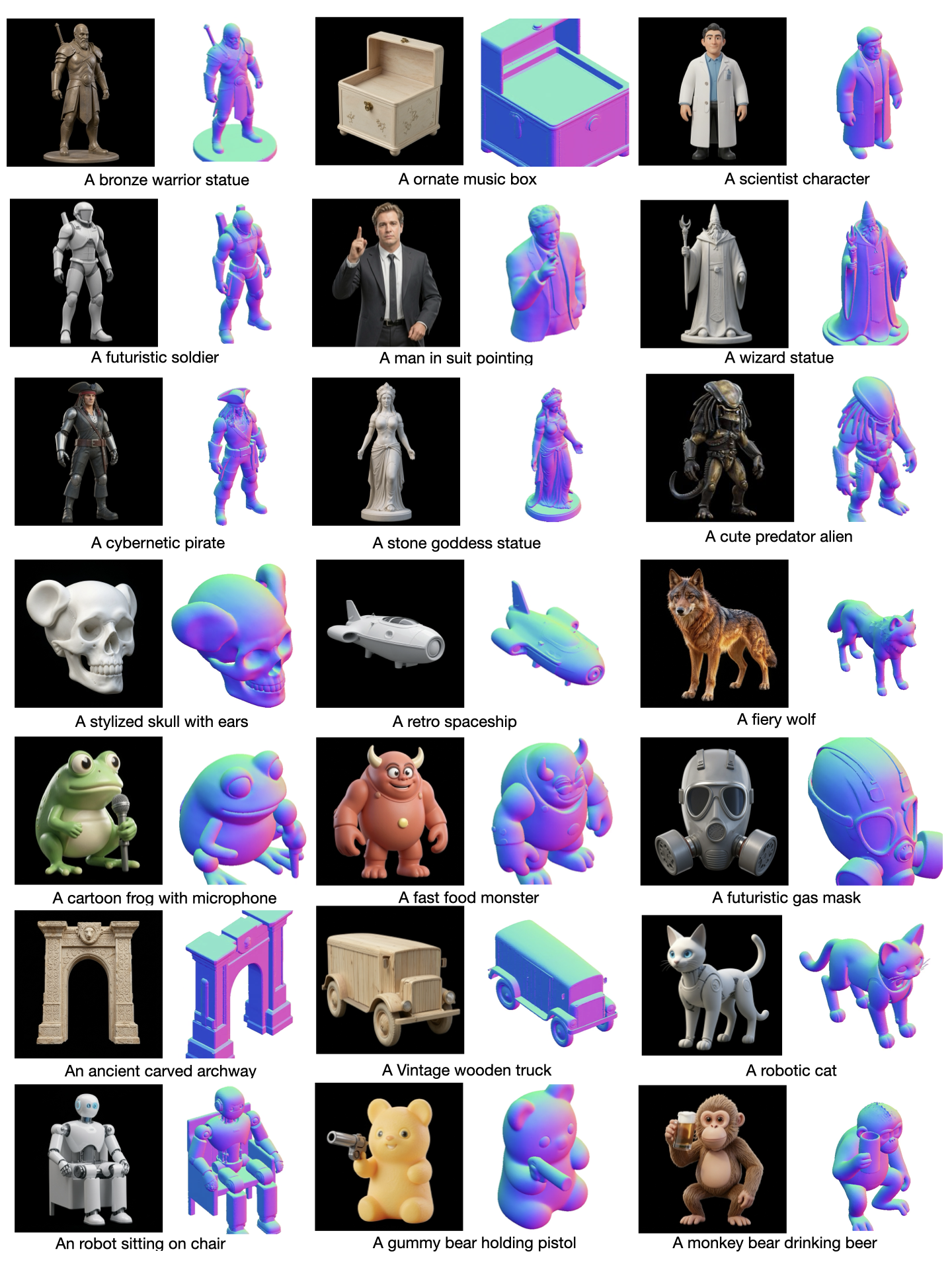}
  \caption{\textbf{Additional qualitative showcase I.} More examples of joint image-and-3D generation by \methodname{}.}
\label{fig:showcase_1}
\end{figure}

\clearpage
\begin{figure}[p]
  \centering
  \includegraphics[width=\textwidth,height=0.9\textheight,keepaspectratio]{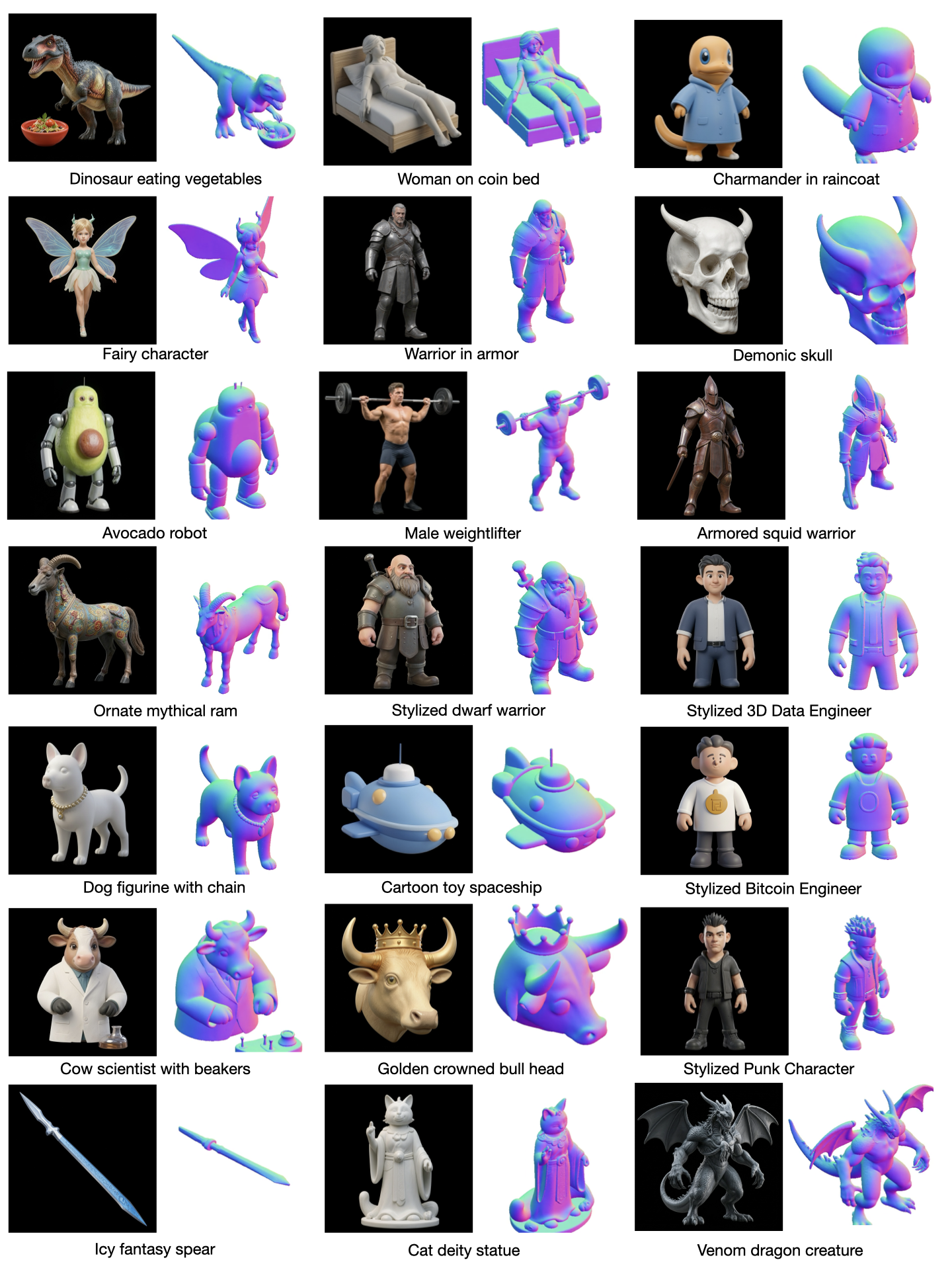}
  \caption{\textbf{Additional qualitative showcase II.} More examples of joint image-and-3D generation by \methodname{}.}
   \label{fig:showcase_2}
\end{figure}

\clearpage

\renewcommand{\refname}{{\color{MeshyLimeDark}References}}






\renewcommand{\refname}{{\color{MeshyLimeDark}References}}

\bibliographystyle{plainnat}
\bibliography{references}

\clearpage
\end{document}